\documentclass[aps,pre,twocolumn,superscriptaddress,showpacs]{revtex4-2}
\usepackage{graphicx}
\usepackage{array}
\usepackage{amsfonts}
\usepackage{amssymb,amsmath,multirow,rotate}
\usepackage{mathrsfs}
\usepackage{booktabs}
\usepackage{threeparttable}
\usepackage{multirow}
\usepackage{epsfig}
\usepackage{threeparttable}
\usepackage{chngpage}
\usepackage{latexsym}
\usepackage{dcolumn}
\usepackage{graphics,graphicx,bm,fleqn,epic,eepic,float}
\usepackage{verbatim}
\usepackage[dvipsnames]{xcolor}
\usepackage{xr}
\usepackage{float}
\usepackage{listings} 
\usepackage{placeins} 
\usepackage{tikz}
\usepackage{amsmath, nccmath}
\usepackage{soul}

\definecolor{red}{rgb}{1,0,0}

\definecolor{blue}{rgb}{0,0,1}

\usepackage{color}
\usepackage[normalem]{ulem}
\definecolor{darkgreen}{rgb}{0.1,.6,.1}

\begin{document}

\title{Vestibular reservoir computing}
\date{\today}

\author{Smita Deb} \email{The first two authors contribute equally} 
\affiliation{School of Electrical, Computer, and Energy Engineering, Arizona State University, Tempe, AZ 85287, USA}

\author{Shirin Panahi} \email{The first two authors contribute equally} 
\affiliation{Department of Electrical and Computer Engineering, Colorado State University, Fort Collins, CO 80523, USA}

\author{Mulugeta Haile}
\affiliation{DEVCOM Army Research Office, 6340 Rodman Road, Aberdeen Proving Ground, MD 21005-5069, USA}

\author{Ying-Cheng Lai} \email{Ying-Cheng.Lai@asu.edu}
\affiliation{School of Electrical, Computer, and Energy Engineering, Arizona State University, Tempe, AZ 85287, USA}
\affiliation{Department of Physics, Arizona State University, Tempe, Arizona 85287, USA}

\begin{abstract}

Reservoir computing (RC) is a computational framework known for its training efficiency, making it ideal for physical hardware implementations. However, realizing the complex interconnectivity of traditional reservoirs in physical systems remains a significant challenge. This paper proposes a physical RC scheme inspired by the biological vestibular system. To overcome hardware complexity, we introduce a designed uncoupled topology and demonstrate that it achieves performance comparable to fully coupled networks. We theoretically analyze the difference between these topologies by deriving a memory capacity formula for linear reservoirs, identifying specific conditions where both configurations yield equivalent memory. These analytical results are demonstrated to approximately hold for nonlinear reservoir systems. Furthermore, we systematically examine the impact of reservoir size on predictive statistics and memory capacity. Our findings suggest that uncoupled reservoir architectures offer a mathematically sound and practically feasible pathway for efficient physical reservoir computing.

\end{abstract}

\maketitle

\section{Introduction} \label{sec:intro}

Reservoir computing (RC), introduced two decades ago~\cite{MNM:2002,jaeger2004,lukovsevivcius2009,MJ:2013}, is a machine learning framework well-suited for dynamics-oriented tasks, most notably nonlinear time-series forecasting~\cite{PLHGO:2017,PHGLO:2018,JL:2019,KKGGM:2020,PCGPO:2021,Bollt:2021,GBGB:2021,Parlitz2024}. Originally proposed to circumvent the training difficulties associated with deep recurrent neural networks~\cite{VPPJHBSTEOKP:2020}, reservoir computing offers distinct advantages, including rapid and stable training, efficient multitasking, and robust performance in sequential learning scenarios. A reservoir computer comprises a network of artificial, typically nonlinear neurons organized in a single hidden layer, referred to as the reservoir network~\cite{DVBSMDDS:2007}. This network maps low-dimensional input signals into a high-dimensional state space, evolves them through nonlinear dynamics, and projects the resulting states back to a low-dimensional output space via a trained readout matrix. For reasonably sized reservoirs, training is computationally efficient, as only the output weights require optimization via standard linear regression. Algorithmically, RC may be realized in either discrete time (updating states via nonlinear activation functions like the hyperbolic tangent) or continuous time (evolving according to differential equations).

Equating the dimensionality of the output vector with that of the input carries significant implications, particularly for applications involving dynamical systems. Upon completion of training, the output vector can be recursively fed back into the input, enabling the reservoir to function as a self-evolving dynamical system. In this manner, a reservoir computer can replicate the ``dynamical climate" of a target system corresponding to the parameters used during training. A further noteworthy capability arises when an additional parameter channel is incorporated to provide explicit, parameter-dependent input to the reservoir. Training on time-series data from a limited set of parameter values enables the reservoir to learn the dependence of the system’s dynamics on that parameter, permitting accurate extrapolation to regions of parameter space where no training data exists. This extrapolative capability positions reservoir computing as a powerful framework for constructing digital twins of nonlinear dynamical systems across broad parameter regimes, facilitating complex tasks such as anticipating critical transitions~\cite{KFGL:2021a,KFGL:2021b} and tipping points~\cite{PKMZGHL:2024}.

Generally, the nonlinear response of a reservoir network to external input must satisfy the \textit{echo state property}. This property imposes several operational criteria: (1) the reservoir dynamics must remain bounded to ensure stability; (2) the input mechanism must map low-dimensional inputs into a high-dimensional state vector, which a linear readout layer then projects back to the output space; (3) the reservoir must exhibit the \textit{fading memory property}, ensuring that the state at time $t$ depends predominantly on recent rather than distant inputs; and (4) in the absence of input, the autonomous dynamics must be independent of initial conditions, ensuring the system ``forgets'' its starting state. Collectively, these conditions delineate the requirements for an effective reservoir. Consequently, any physical system meeting these criteria can, in principle, serve as a reservoir, a concept central to the emerging field of physical reservoir computing~\cite{TYHNKTNNH:2019,NK:2020}.

The significance of physical reservoir computing lies in its capacity to reduce reliance on digital computation by exploiting the intrinsic dynamical properties of physical systems for information processing~\cite{NKFI:2021}. This paradigm offers advantages in energy efficiency, latency reduction, scalability, and security, presenting a promising alternative to conventional architectures~\cite{TYHNKTNNH:2019,NK:2020}. By leveraging naturally occurring nonlinearities and high-dimensional transformations, physical RC enables analog computation that integrates seamlessly with real-world environments while minimizing dependence on digital processors~\cite{stepney2024,Zhang2023}. Over the past decade, the concept has been realized across diverse platforms, including electronic~\cite{appeltant2011,moran2023,milano2022,zhong2022,rajib2022,liang2024}, photonic~\cite{LBMUCJ:2017,du2022,picco2024}, mechanical~\cite{wang2024}, and quantum systems~\cite{abbas2024,palacios2024,Zhu2025}. More recently, efforts to exploit living organisms and biological mechanisms as computational substrates have further expanded the scope of the field~\cite{Soriano2015,KKGGM:2020,Cucchi2021,Illeperuma2024,He2025}. These advances underscore the potential of physical reservoir computing as a foundation for next-generation analog and hybrid computational technologies.

While physical realizations of RC hold promise for bridging computational theory and hardware-based processing, their implementation presents inherent challenges. A primary difficulty lies in constructing large networks of physical nodes to sufficiently expand the low-dimensional input into a high-dimensional state space~\cite{cucchi2022}. A second challenge concerns the realization of appropriate nodal coupling schemes. In software-based implementations, large randomly initialized recurrent neural networks with complex topologies are standard~\cite{Schrauwen2007,Ma2023,gonon2024}. In contrast, physical implementations rely on alternative techniques, each with limitations. An influential approach is the single-node, time-delayed feedback reservoir~\cite{HSRFG:2015,dion2018}, where a single nonlinear node with a feedback loop generates a high-dimensional representation via temporal multiplexing. This configuration underlies various photonic and electronic systems~\cite{li2017,Tran2020,Boshgazi2022}. Other methods utilize cellular automata~\cite{nichele2017} or networks of coupled oscillators (e.g., DNA oscillators~\cite{goudarzi2013}). Despite their success, these implementations often demand intricate hardware configurations requiring precise calibration.

For practical realization, compact reservoir networks with simple coupling architectures are highly desirable. The simplest configuration is the \textit{uncoupled reservoir}, where nodes evolve independently. Recent studies~\cite{Jaurigue2024} demonstrate that such uncoupled reservoirs—comprising independent dynamical elements processing signals in parallel—can effectively perform tasks like chaotic time-series prediction. The absence of inter-node coupling eliminates the need for fine-tuning interaction parameters and facilitates scalability. However, an outstanding conceptual question remains: why do uncoupled networks perform effective computation when the echo state property typically relies on recurrent connectivity for dynamical memory? Understanding how memory and computational capability emerge in uncoupled architectures remains a critical open issue.

In this paper, we propose \textit{vestibular reservoir computing}, a biologically inspired framework modeled after the vestibular system and implemented via an uncoupled network architecture. The vestibular system, located in the inner ear, is a sophisticated sensory organ responsible for balance, posture, and motion detection [see Supplementary Information (SI), Sec.~\ref{secA1}]. Its ability to integrate diverse mechanical and neural signals makes it an inspiring template for physical RC. We develop a simplified mathematical model capturing the essential functionalities of the vestibular system, incorporating the mechanical dynamics of semicircular canals and otolith organs alongside the neural response of hair cells, while remaining computationally tractable. We implement vestibular reservoir computing in both uncoupled and coupled architectures for comparative evaluation. Remarkably, the uncoupled configuration achieves performance comparable to, and occasionally exceeding, the coupled network in predicting complex dynamical systems. To investigate the underlying mechanism, we examine \textit{memory capacity}, a metric quantifying the system's ability to reconstruct past inputs. We theoretically establish conditions under which the memory capacity of a linear uncoupled reservoir equals that of a coupled one, extending this relationship approximately to nonlinear systems. Numerical simulations confirm these conditions, explaining the robust performance of our vestibular model. These results demonstrate that biologically inspired, uncoupled architectures can achieve high efficiency and memory performance, paving the way for low-complexity physical reservoir computing systems.

\section{Results} \label{sec:results}
 
\begin{figure} [ht!]
\centering
\includegraphics[width=\linewidth]{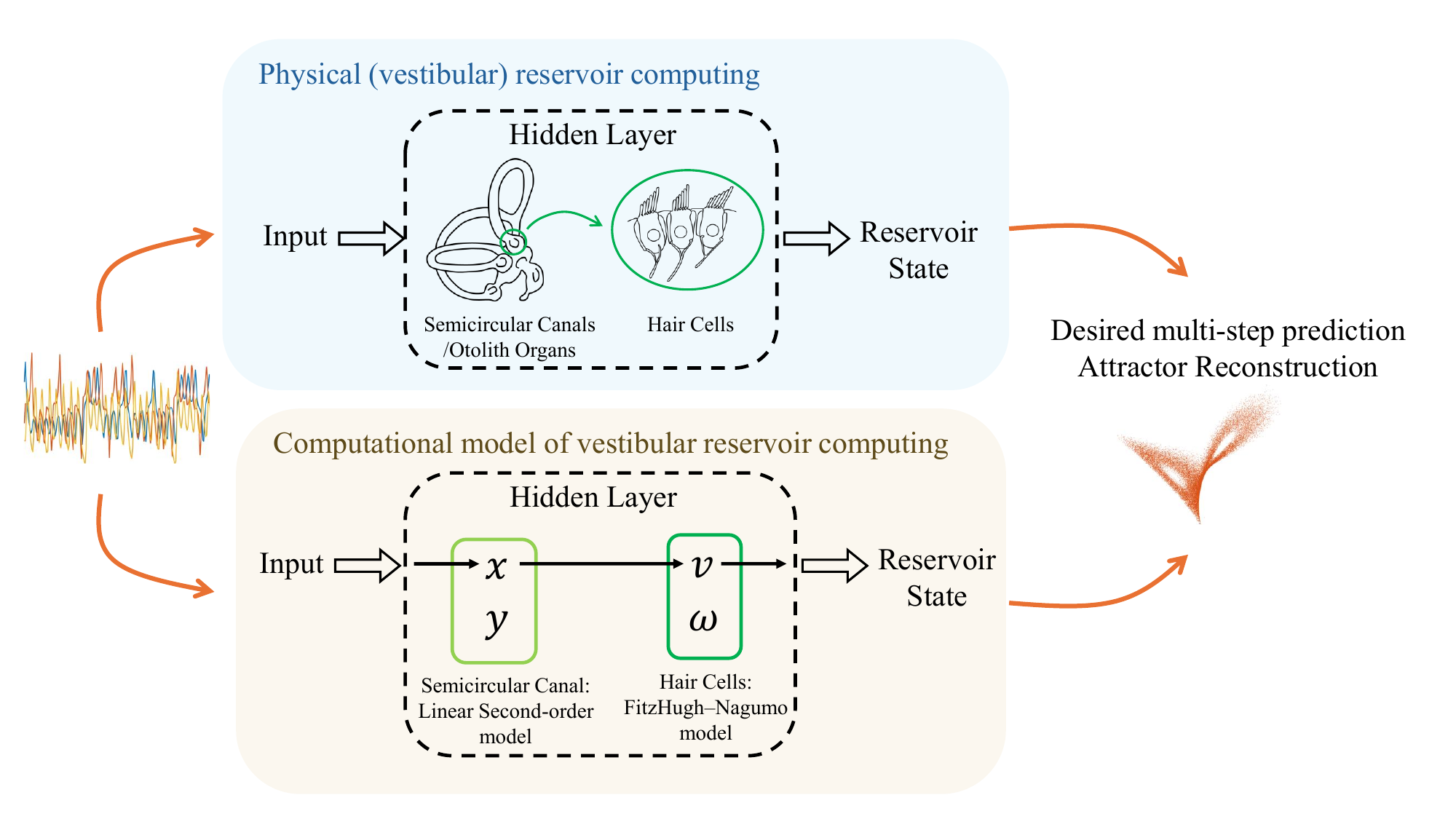}
\caption{Schematic representation of the proposed vestibular reservoir computing (RC) scheme. The reservoir incorporates biomechanical structures (semicircular canals and otolith organs) that process external stimuli via endolymph movement, alongside neurophysiological components (hair cells) that transduce mechanical stimuli into neural signals. The computational framework models these interactions by coupling semicircular canal mechanics with FitzHugh–Nagumo (FHN) neural dynamics.}
\label{fig:fig1}
\end{figure}

To realize vestibular reservoir computing, we begin by analyzing basic vestibular dynamics through a mathematical modeling approach that captures essential functions while maintaining computational simplicity. Rather than developing a highly detailed biophysical model, our objective is to construct a generalized mathematical representation that provides insight into the holistic behavior of the vestibular system. This abstraction emphasizes the system-level interactions between biomechanical and neurophysiological components underlying vestibular processing, rather than specific physiological minutiae. Such a formulation provides a flexible foundation for integrating vestibular dynamics into physical reservoir computing. As shown in Fig.~\ref{fig:fig1}, a vestibular reservoir computer consists of two key components. The first component utilizes a second-order linear approximation for the semicircular canals and otolith organs to describe the fluid dynamics and mechanical properties of the system. The second component is a two-dimensional nonlinear FitzHugh–Nagumo (FHN) neuron model, which approximates the neural signal processing occurring in hair cells.

The mathematical formulation providing a coupled representation of mechanical dynamics and neural activity in the vestibular system is given by: 
\begin{subequations} \label{SCCFHN}
\begin{alignat}{5}
\dot{\mathbf{x}} &=\mathbf{ y},\\ 
\dot{\mathbf{y}} &= \frac{1}{m}(-c \mathbf{y} - k \mathbf{x}) +  (\mathbb{A}\cdot\mathbf{x} + \mathbb{W}_{\rm in} \cdot\mathbf{u}), \\
\dot{\mathbf{v}} &= (d \mathbf{v} - \frac{\mathbf{v}^{\circ 3}}{3} - \boldsymbol{\omega}) + \mathbf{I}_{ext},\\
\dot{\boldsymbol{\omega}} &= \mathbf{v} + a - b\boldsymbol{\omega}
\end{alignat}
\end{subequations}
where $\mathbf{x}$ and $\mathbf{y}$ represent the displaced volume and speed of the endolymph in the semicircular canals, respectively; $m$ is the mass of the fluid contained in the canal; $c$ describes the damping effect of the viscous fluid; and $k$ represents the stiffness of the cupula. The variable $\mathbf{v}$ represents the neuronal membrane voltage, $\circ$ denotes the Hadamard power, and $\boldsymbol{\omega}$ is a slower linear recovery variable modeling the reactivation and deactivation of sodium and potassium channels following stimulation. The external input current derived from the vestibular stimulus is given by $\mathbf{I}_{\rm ext} = \sigma \mathbf{x}$, where $\sigma$ is a scaling factor. The reservoir connectivity matrix is denoted by $\mathbb{A}$, $\mathbf{u}$ is the $D$-dimensional input vector, and $\mathbb{W}_{\rm in}$ is the input weight matrix. Further details regarding reservoir parameters are presented in the subsequent section. For this study, the vestibular system parameters are set as follows: $c=12$, $m=2$, $k=50$, $d=-3.8$, $\sigma=6.5$, $a=0.7$, and $b=2$.

To evaluate the performance of the vestibular reservoir computer, we consider a simplified configuration in which the FHN neuron model parameters are selected to produce a fixed-point steady state, thereby satisfying the echo state property. For a systematic comparative evaluation, we first examine the conventional reservoir computing setup, where the reservoir consists of a coupled network with a random sparse topology characterized by a random connection matrix $\mathbb{A}$ in Eq.~\eqref{SCCFHN}. We then focus on uncoupled reservoir networks, where $\mathbb{A}$ is diagonal. All numerical experiments follow a structured protocol comprising training, validation, and testing phases. The training and validation phases constitute the open-loop process, while the testing phase functions as the closed-loop process (see Methods~\ref{M_model} and \ref{M_RC}).

\subsection{Vestibular reservoir computing with coupled network}  

In the coupled configuration, the reservoir network consists of interconnected nodes defined by $\mathbb{A}$, a symmetric, random sparse matrix with elements drawn independently from a uniform distribution. The link density is set to $0.4$; thus, for a network of size $N$, approximately $0.4N^2$ elements in $\mathbb{A}$ are nonzero. Subsequently, $\mathbb{A}$ is rescaled such that the resulting matrix possesses negative eigenvalues and a prescribed spectral radius $\rho$ (the magnitude of the largest eigenvalue). This spectral radius plays a critical role in maintaining the overall stability and memory capacity of the reservoir. The input weight matrix $\mathbb{W}_{\text{in}} \in \mathbb{R}^{N \times D}$ is fixed, with elements uniformly distributed in the range $[-\gamma,\gamma]$, where $D$ denotes the input dimension.

\begin{figure*} [ht!]
\centering
\includegraphics[width=1\linewidth]{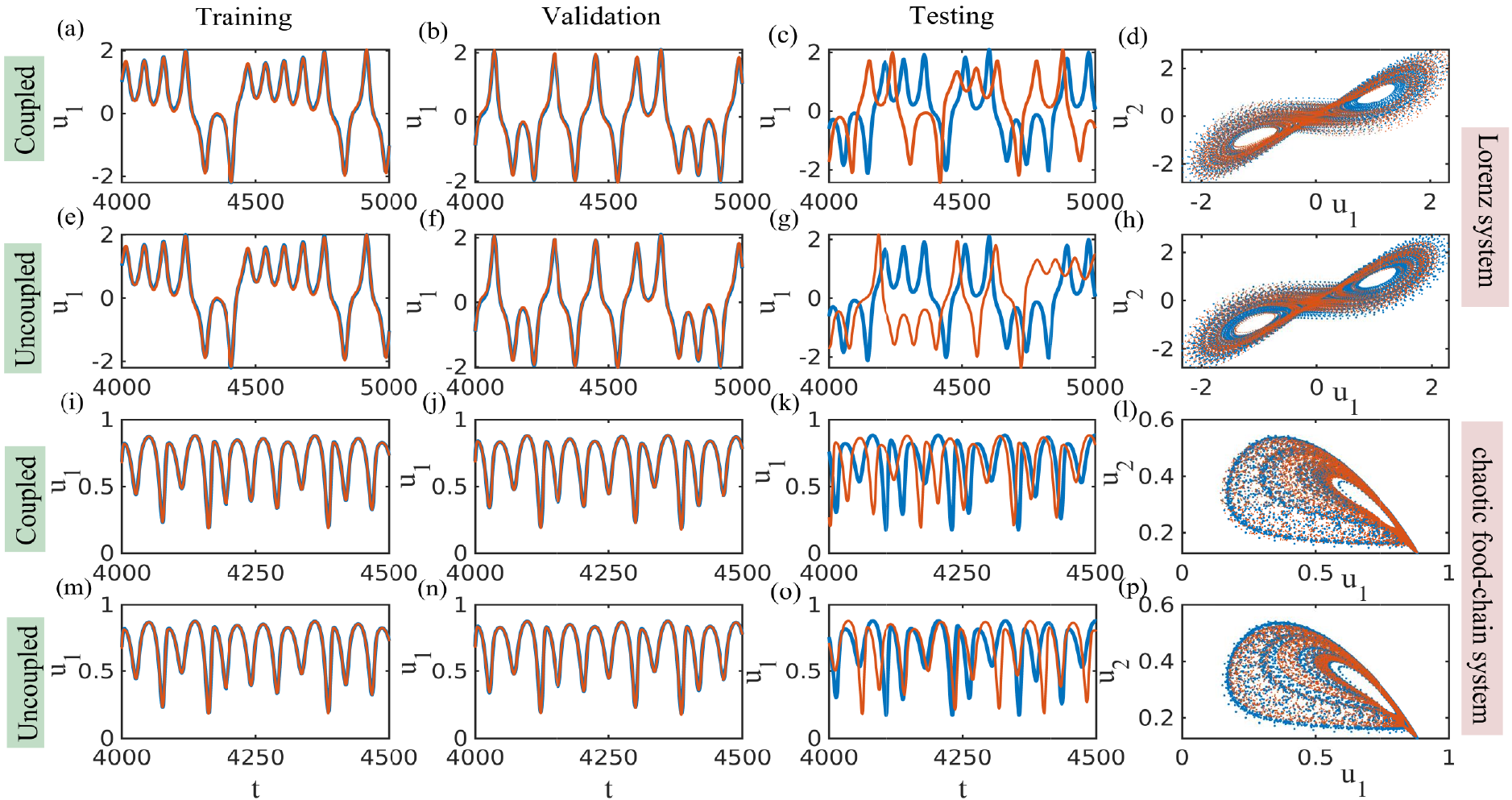}
\caption{Performance comparison of coupled and uncoupled vestibular reservoir computers for Lorenz and chaotic food-chain systems. (a-d) Results for the coupled configuration trained on the Lorenz system, showing training, validation, closed-loop testing, and closed-loop attractor reconstruction, respectively. The validation phase evaluates generalization on unseen data in an open-loop setup, while attractor reconstruction demonstrates the autonomous system's ability to reproduce long-term chaotic dynamics. (e-h) Corresponding results for the uncoupled vestibular reservoir computer trained on the Lorenz system. (i-l) Results for the coupled configuration trained on the chaotic food-chain system. (m-p) Corresponding results for the uncoupled configuration on the food-chain system. The time series plots display the variables $u_{1}$ and $u_{2}$ from Eq.~\eqref{eq:eqinput1} and Eq.~\eqref{eq:eqinput2}. Hyperparameter values are listed in SI Sec.~\ref{secH}, Tab.~\ref{tab:example}.}
\label{fig:fig2}
\end{figure*}

Figure~\ref{fig:fig2} illustrates the short-term and long-term prediction performance of the vestibular reservoir computer, highlighting the system's ability to learn and forecast complex time-series data. Specifically, the results in the first and third rows display the performance of a coupled $30$-node vestibular reservoir computer trained on the chaotic Lorenz and food-chain systems, respectively. For the Lorenz system, the training and validation errors are $0.013$ and $0.015$, respectively. Similarly, the errors for the chaotic food-chain system are $0.006$ (training) and $0.007$ (validation). These low error values confirm that the system effectively learns the chaotic dynamics in an open-loop configuration, achieving accurate one-step-ahead predictions.

In the testing phase, the reservoir computer operates autonomously. We observe that multi-step predictive performance remains stable over an extended time horizon. As shown in Figs.~\ref{fig:fig2}(d) and \ref{fig:fig2}(l), the reconstructed trajectories from the autonomous vestibular reservoir computer remain confined within the Lorenz and food-chain attractors, respectively. For the Lorenz and food-chain systems, the deviation values (DV) are $0.330$ and $0.364$, respectively, indicating statistically that the system preserves the underlying strange attractor structure of the original systems while maintaining excellent long-term prediction capability. To further quantify the divergence between the ground truth and the prediction, we calculate the Kullback-Leibler (KL) divergence, which yields values of $0.0006$ for the Lorenz system and $0.0007$ for the food-chain system. Additionally, the largest Lyapunov exponents of the attractors generated by the vestibular reservoir computer are $0.030$ (Lorenz) and $0.021$ (food-chain), matching closely with the ground truth values of $0.03$ and $0.023$, respectively. For further details regarding the calculation of these statistical properties, see SI Sec.~\ref{si_statistic}.

Regarding physical implementation, a potential approach for realizing the coupling matrix $\mathbb{A}$ in vestibular reservoir computing involves a ferromagnetic fluid system, where the movement of one element influences the magnetic and electric fields of neighboring nodes. However, such implementations often suffer from low accuracy and practical physical constraints, rendering them challenging for high-performance applications. These limitations underscore the necessity of investigating the performance of uncoupled reservoir network configurations.

\begin{figure*} [ht!]
\centering
\includegraphics[width=0.8\linewidth]{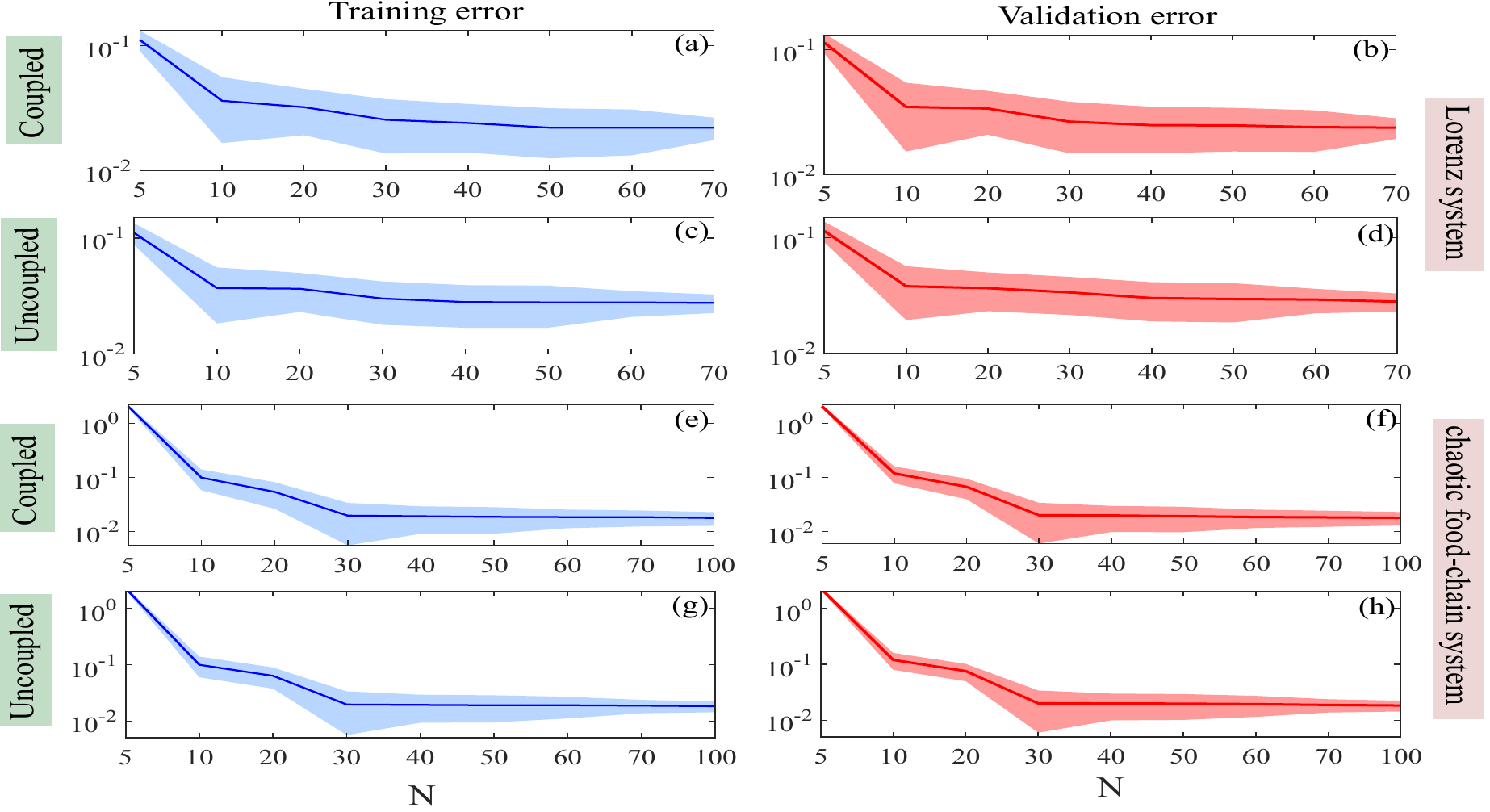}
\caption{Short-term predictive performance of coupled and uncoupled vestibular reservoir computers. The plots show the mean training and validation errors (quantified as Normalized Root-Mean-Square Error, NRMSE) as a function of reservoir network size for the (a-d) Lorenz and (e-h) chaotic food-chain systems. The shaded regions represent the standard deviation on a logarithmic scale, calculated over $100$ independent trials.}
\label{fig:error}
\end{figure*}
	
\begin{figure*} [ht!]
\centering
\includegraphics[width=0.8\linewidth]{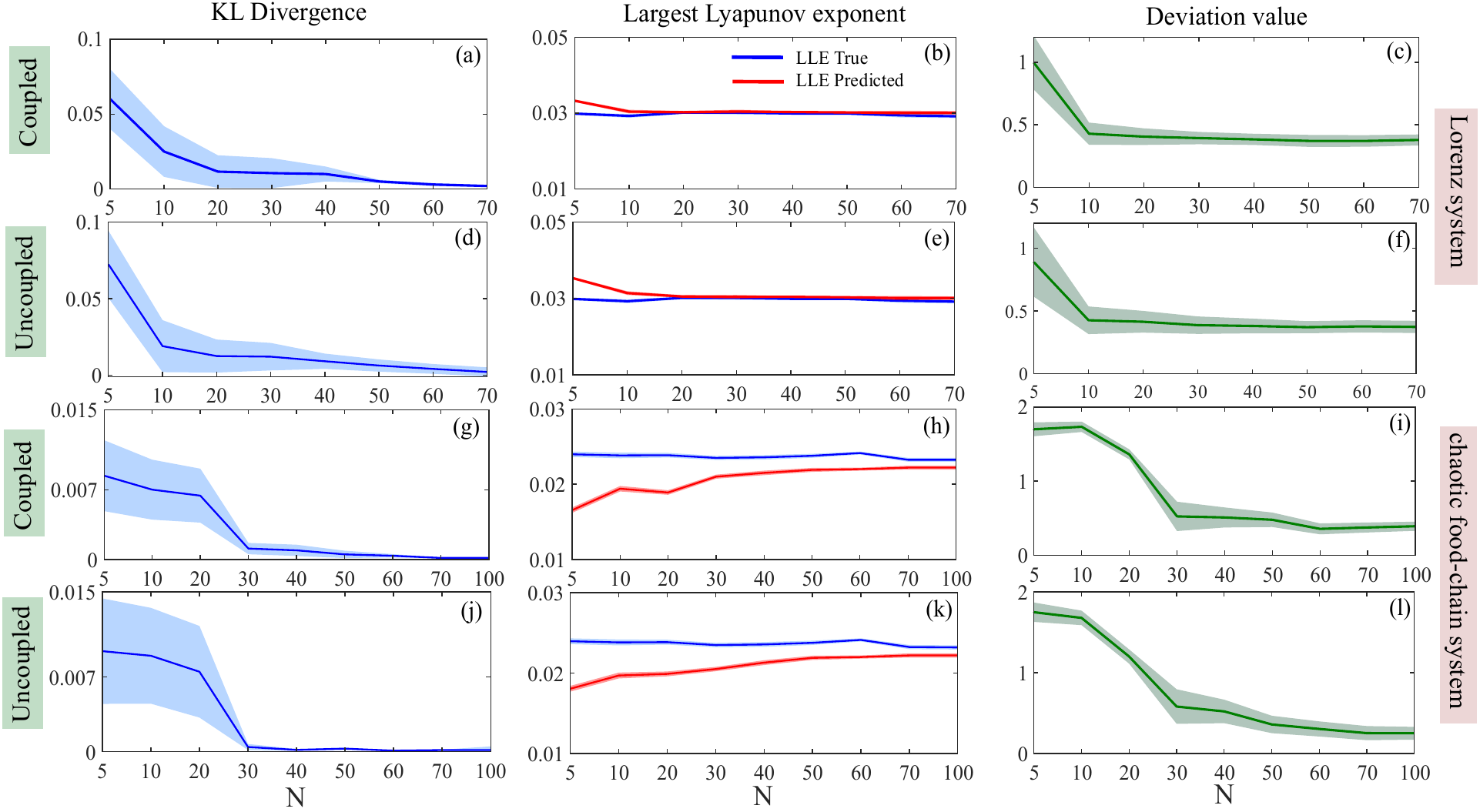}
\caption{Predictive statistics of coupled and uncoupled vestibular reservoir computers. The plots display the Kullback-Leibler (KL) divergence, largest Lyapunov exponent, and mean Deviation Value (DV) as a function of network size. Panels correspond to: (a-c) coupled reservoir on the Lorenz system; (d-f) uncoupled reservoir on the Lorenz system; (g-i) coupled reservoir on the food-chain system; and (j-l) uncoupled reservoir on the food-chain system. Shaded regions indicate the standard deviation calculated on a linear scale from 100 independent trials.}
\label{fig:error1}
\end{figure*}

\subsection{Vestibular reservoir computing with uncoupled network}

For an uncoupled reservoir network, the connectivity matrix $\mathbb{A}$ in Eq.~\eqref{SCCFHN} is a diagonal matrix. All other aspects of the reservoir configuration, including the nonlinear dynamical equations and the training procedure, remain unchanged. The second and fourth rows in Fig.~\ref{fig:fig2} illustrate the predictive performance for Lorenz and chaotic food-chain inputs, respectively, where the reservoir network comprises $30$ nodes—or, equivalently, $30$ independent, parallel processing units. The open-loop training and validation (one-step prediction) errors are $0.018$ and $0.019$, respectively, for the Lorenz system; the corresponding errors for the food-chain system are $0.009$ and $0.009$. These low error values indicate high predictive accuracy.

For multi-step prediction in the testing phase, the autonomous system generates trajectories that remain bounded within the Lorenz and food-chain attractors, as shown in Figs.~\ref{fig:fig2}(h) and \ref{fig:fig2}(p), respectively. (Figure~\ref{fig:figsi31} in the SI presents the results of closed-loop attractor reconstruction for all variables using predictions from both coupled and uncoupled vestibular reservoir computing for these systems.) The estimated largest Lyapunov exponents of the uncoupled vestibular reservoir computer for the target Lorenz and chaotic food-chain systems are $0.030$ and $0.021$, which are very close to the respective ground truth values of $0.030$ and $0.023$. The corresponding DV values are $0.318$ and $0.355$, and the KL divergence values are $0.0027$ and $0.0004$ for the Lorenz and chaotic food-chain systems, respectively. These metrics confirm that the reservoir computer reproduces the essential dynamics over long-term evolution. Collectively, these results demonstrate that the uncoupled vestibular reservoir computer offers a viable alternative to its coupled counterpart, achieving comparable performance while offering significant advantages in physical implementation.

Specifically, we use 100 realizations to identify the frequency with which the reservoir computer generates an attractor that remains close to the ground truth. For both uncoupled and coupled vestibular reservoir computers, no divergent cases are observed when the network contains more than $30$ nodes. As the network size falls below $30$, the number of divergent cases begins to increase. In these instances, we calculate the RMSE for short-term time-series prediction, as well as the KL divergence, largest Lyapunov exponent, and DV values for long-term attractor prediction, considering only the performing (non-divergent) cases.

\subsection{How small can an uncoupled vestibular reservoir computer be?}

For practical implementation, minimizing reservoir network size is highly desirable. In the context of vestibular reservoir computing, we aim to determine the minimal network size required to deliver reasonable predictive performance. To systematically evaluate the impact of network size, we decrease the number of nodes in the uncoupled reservoir network and analyze the resulting predictive accuracy in comparison with the coupled counterpart. generally, as the reservoir network becomes smaller, the probability of prediction failure increases. It is therefore useful to quantify how this probability evolves as the network size decreases. We estimate this probability by employing a large number of independent realizations of the closed-loop operation.

Short-term and long-term performances are illustrated in Figs.~\ref{fig:error} and \ref{fig:error1}, respectively. Specifically, Fig.~\ref{fig:error} displays the training and validation errors across different reservoir sizes. As anticipated, larger reservoirs exhibit lower training and validation errors due to their increased capacity to capture complex temporal dependencies in the input data; this trend holds for both uncoupled and coupled architectures. Figure~\ref{fig:error1} presents a detailed statistical analysis of long-term predictive performance. It is evident that, provided the reservoir network exceeds 30 nodes, the KL divergence and DV values remain small, and the largest Lyapunov exponent converges to the ground truth value. These results highlight a trade-off between reducing reservoir size and maintaining prediction accuracy. While increasing the node count enhances both short- and long-term performance, a moderate reduction in reservoir size can still yield satisfactory accuracy, thereby facilitating physical implementations with reduced computational and hardware demands. Overall, the predictive trends remain consistent across both coupled and uncoupled vestibular reservoir computing architectures.

\subsection{Memory function and memory capacity of a reservoir computer}

\begin{figure*} [ht!]
\includegraphics[width=0.8\linewidth]{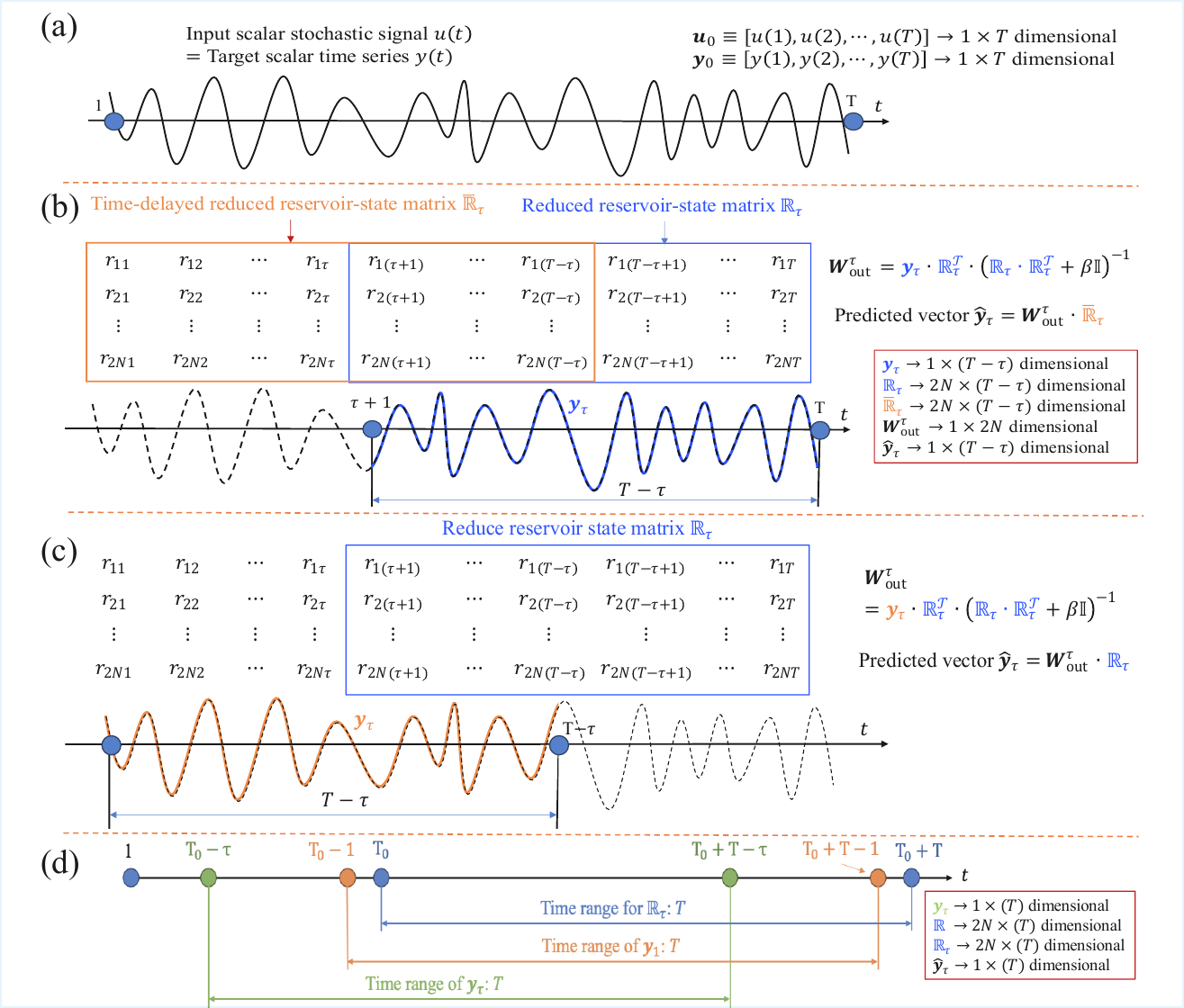}
\caption{Methods for calculating the memory function. (a) A scalar stochastic signal $u(t)$ serves as both the input and the target signal $y(t)$ for training. (b) Schematic of the standard calculation method. The full reservoir state matrix $\mathbb{R}$ has dimensions $2N\times T$. The reduced state matrix $\mathbb{R}_{\tau}$ (blue) is defined over the interval $[\tau + 1, T]$, while the time-delayed state matrix $\bar{\mathbb{R}}_{\tau}$ covers $[1,T-\tau]$. The output weight vector $\mathbf{W}_{\rm out}^{\tau}$ is trained using $\mathbb{R}_{\tau}$ and the target series $\mathbf{y}_{\tau}$ (interval $[\tau + 1, T]$). The predicted series $\hat{\mathbf{y}}_{\tau}$ is then computed as the product of $\mathbf{W}_{\rm out}^{\tau}$ and the time-delayed states $\bar{\mathbb{R}}_{\tau}$. (c) Alternative calculation method. Here, $\mathbf{W}_{\rm out}^{\tau}$ is trained using the time-delayed target series $\mathbf{y}_{\tau}$ and the reduced state matrix $\mathbb{R}_{\tau}$. The prediction $\hat{\mathbf{y}}_{\tau}$ is obtained by multiplying $\mathbf{W}_{\rm out}^{\tau}$ and $\mathbb{R}_{\tau}$ directly. (d) The refined method used to eliminate statistical imbalance (see text). The relevant time intervals are adjusted such that the reservoir state definition remains constant ($\mathbb{R}_{\tau}\equiv \mathbb{R}$) for all $\tau$.}
\label{fig:MC}
\end{figure*}

Why does the uncoupled vestibular reservoir computer deliver prediction performance comparable to its coupled counterpart? To gain insight into this phenomenon, we examine a fundamental characteristic of reservoir computing: \textit{memory capacity}, defined as 
the ``system’s ability to reconstruct past inputs”~\cite{jaeger2001short,carroll2022optimizing,dambre2012information}. Since memory capacity is an intrinsic property of the reservoir that should be independent of the correlations typically present in chaotic inputs, it is necessary to utilize a stochastic input signal, denoted as $u(t)$ [see Fig.~\ref{fig:MC}(a); note that the signal appears smooth only for visualization purposes]. The target signal used to train the reservoir is $y(t) = u(t)$. Assuming successful training, the output signal $\hat{y}(t)$ should closely approximate the target: $\hat{y}(t) \approx y(t)$. In this context, memory capacity quantifies the reservoir's ability to reconstruct past inputs based on its current state.

\subsubsection{Analysis of memory capacity for linear reservoir networks}

To quantitatively define memory capacity, we recall that the training objective is to fit an output matrix via ridge regression using the target signal. This matrix is then used to compute the predicted signal [see Eqs.~\eqref{eq:RC_Ridge_Regression} and (\ref{eq:RC_Training_Output}) in \textbf{Methods}]. All information regarding the target signal within the time interval $[0,T]$ is embedded in the $2N\times T$-dimensional reservoir state matrix $\mathbb{R}$, where $T$ is sufficiently large to mark the end of training. While the reservoir consists of $N$ nodes, the reservoir states are augmented by including their squared values to avoid mirror attractors, thereby increasing the effective dimensionality to $2N$. Reconstructing the ``past" implies using the current state to predict a time-delayed version of the signal. To implement this, we define a reduced reservoir state matrix $\mathbb{R}_{\tau}$ with dimension $2N\times (T-\tau)$, as illustrated in Fig.~\ref{fig:MC}(b). The corresponding reservoir output matrix $\mathbf{W}^{\tau}_{\rm out}$ is obtained using the delayed target time series $\mathbf{y}_{\tau} = \mathbf{u}_{\tau}$, restricted to the interval $[\tau+1,T]$. By applying the $\tau$-delayed reduced reservoir state matrix $\bar{\mathbb{R}}_{\tau}$, we generate the prediction $\hat{\mathbf{y}}_{\tau} = \mathbf{W}^{\tau}_{\rm out}\cdot \bar{\mathbb{R}}_{\tau}$, corresponding to the $\tau$-delayed ``past" of the target signal [Fig.~\ref{fig:MC}(b)]. The squared linear correlation between the reduced target series $\mathbf{y}_{\tau} = \mathbf{u}_{\tau}$ and the predicted series $\hat{\mathbf{y}}_{\tau}$ defines the memory function: 
\begin{subequations}
\begin{align} \label{eq:MF}
        {\rm M}_F(\tau) &= \frac{\bigl({\rm Cov}[\mathbf{u}_{\tau},\hat{\mathbf{y}}_{\tau}]\bigl)^2}{{\rm Var}[\mathbf{u}_{\tau}]\cdot {\rm Var}[\hat{\mathbf{y}}_{\tau}]} \\ \label{eq:MFa}
        &= \frac{\bigl(\sum^{T-\tau}_{k=1}[u_{\tau}(k) - \bar{u}_{\tau}][\hat{y}_{\tau}(k) - \bar{y}_{\tau}]\bigl)^2}{\bigl(\sum^{T-\tau}_{k=1}[u_{\tau}(k) - \bar{u}_{\tau}]^2\bigl)\bigl(\sum^{T-\tau}_{k=1}[\hat{y}_{\tau}(k) - \bar{y}_{\tau}]^2\bigl)},
\end{align}
\end{subequations}
where $u_{\tau}(k)$ and $\hat{y}_{\tau}(k)$ are the delayed scalar signals. For convenience, the indices of $\mathbf{u}_{\tau}$ and $\mathbf{y}_{\tau}$ are rearranged to run from $1$ to $T-\tau$, with $\bar{u}_{\tau}$ and $\bar{y}_{\tau}$ representing their respective means over this interval. The correlation as a function of $\tau$ characterizes the reservoir's memory capacity~\cite{jaeger2001short,carroll2022optimizing,dambre2012information}; specifically, a slower decay of correlation with increasing $\tau$ indicates higher capacity. This relationship arises because the reduced reservoir state matrix $\mathbb{R}_{\tau}$ overlaps with its time-delayed version $\bar{\mathbb{R}}_{\tau}$ in the interval $[\tau+1,T-\tau]$. This overlapping region contributes to the correlation; as $\tau$ increases, the overlap shrinks, causing the memory function to decrease. The total memory capacity of the reservoir is proportional to the area under the ${\rm M}_F(\tau)$ curve, given by~\cite{jaeger2001short,carroll2022optimizing,dambre2012information}: 
\begin{align} \label{eq:MC} 
	{\rm M}_C \equiv \sum^{T-1}_{\tau = 0} {\rm M}_F(\tau). 
\end{align}
An alternative method for calculating the memory function involves using the target series $\mathbf{y}_{\tau}$ delayed by $\tau$ (from the interval defining $\mathbb{R}_{\tau}$) to calculate the output weight vector $\mathbf{W}_{\rm out}^{\tau}$, as shown in Fig.~\ref{fig:MC}(c), while using the original state matrix $\mathbb{R}_{\tau}$ to obtain the predicted series $\hat{\mathbf{y}}_{\tau}$, rather than the time-delayed matrix $\bar{\mathbb{R}}_{\tau}$ used in Fig.~\ref{fig:MC}(b). However, a drawback of both methods illustrated in Figs.~\ref{fig:MC}(b) and (c) is that the summation range decreases as $\tau$ increases (for a fixed large $T$), potentially introducing statistical imbalance. To remedy this, we designate an intermediate time interval $[1,T_0]$ where $\tau < T_0$, as shown in Fig.~\ref{fig:MC}(d). In this approach, the reservoir state matrix is defined over the interval $[T_0,T_0+T]$ with a fixed duration $T$. The time ranges for sampling the target signal to calculate $\mathbf{W}_{\rm out}^{\tau}$ are also fixed at length $T$ for any $\tau$. This ensures that the reservoir definition remains consistent regardless of delay, allowing for a precise evaluation of how well a delayed input lies within the subspace spanned by the reservoir. The memory function in Eq.~\eqref{eq:MF} then takes the form: 
\begin{align} \label{eq:MF_updated}
{\rm M}_F(\tau) =  \frac{\bigl(\sum^{T}_{k=1}[u_{\tau}(k) - \bar{u}_{\tau}][\hat{y}_{\tau}(k) - \bar{y}_{\tau}]\bigl)^2}{\bigl(\sum^{T}_{k=1}[u_{\tau}(k) - \bar{u}_{\tau}]^2\bigl)\bigl(\sum^{T}_{k=1}[\hat{y}_{\tau}(k) - \bar{y}_{\tau}]^2\bigl)},
\end{align}

\begin{figure*} [ht!]
\centering
\includegraphics[width=0.8\linewidth]{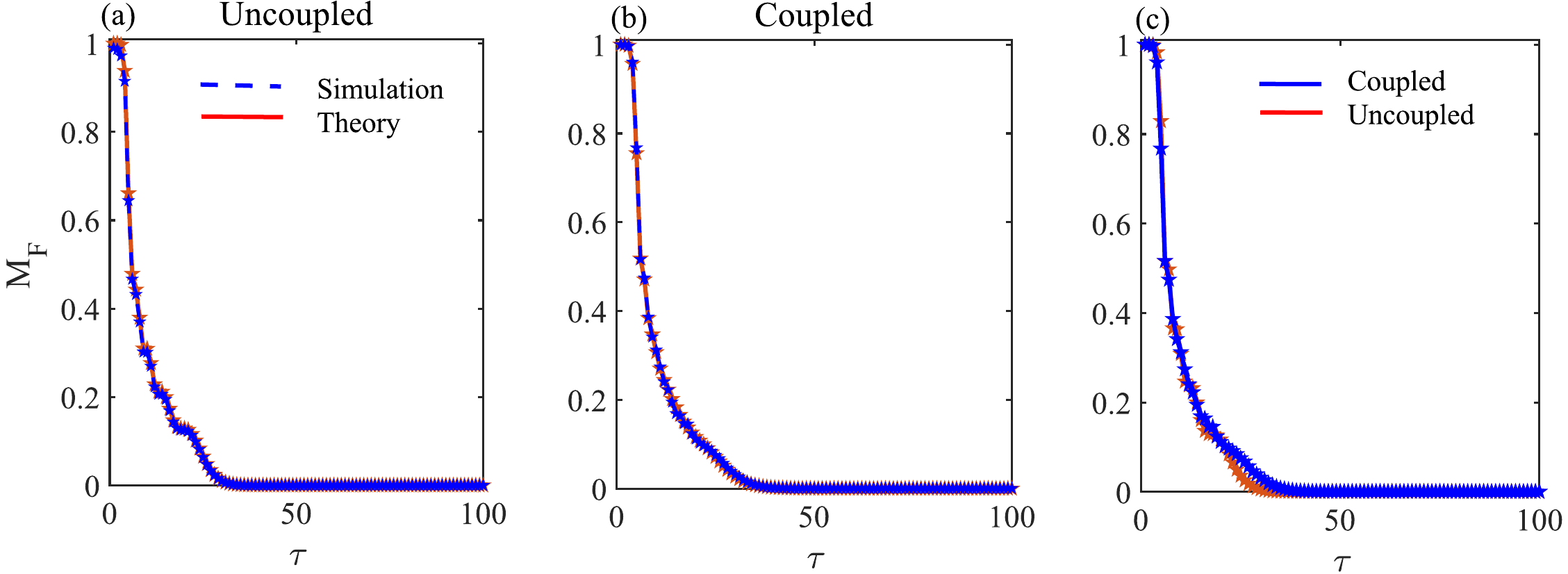}
\caption{Memory function ${\rm M}_F$ for a linear reservoir computer driven by a scalar stochastic input. (a, b) ${\rm M}_F$ versus time delay for an uncoupled and a coupled linear reservoir network, respectively. Orange and blue traces represent theoretical predictions and numerical simulation results, respectively. (c) Numerical computation of ${\rm M}_F$ for an uncoupled linear reservoir (orange) configured to share the same eigenvalue spectrum as the coupled linear reservoir (blue). All simulations are averaged over 100 independent trials with a spectral radius $\rho=0.9$.}
\label{fig:MC_val}
\end{figure*}

To gain analytic insight into the role of information processing capacity in the predictive capability of coupled versus uncoupled reservoir networks, we analyze reservoir networks governed by linear activation functions. The focus of this analysis is to identify the conditions under which an uncoupled reservoir computer exhibits performance comparable to its coupled counterpart. These conditions can subsequently be tested numerically for conventional reservoir computing with nonlinear activation functions.

For a scalar input signal and a scalar output, the reservoir output matrix $\mathbb{W}_{\rm out}$ reduces to a row vector, denoted as $\mathbf{W}_{\rm out}$. The memory function ${\rm M}_F(\tau)$ for a given delay $\tau$ in Eq.~\eqref{eq:MF_updated} can be expressed equivalently in terms of the reservoir readout weight vector $\mathbf{W}_{\rm out}^{\tau}$ obtained by minimizing the mean-squared error loss function. Specifically, an equivalent expression for ${\rm M}_F(\tau)$ is given by~\cite{dambre2012information}:
\begin{align}\label{eq:eqMFN3}
        {\rm M}_F(\tau)=1-\frac{min_{\mathbf{W}_{\rm out}^{\tau}} [(\mathbf{\hat{y}}_{\tau}-\mathbf{u}_{\tau})\cdot(\mathbf{\hat{y}}_{\tau}-\mathbf{u}_{\tau})^{\intercal}]}{\mathbf{u}_{\tau}\cdot\mathbf{u}_{\tau}^{\intercal}}.
\end{align}
Using the relation 
\begin{align} \nonumber
\mathbf{\hat{y}}_{\tau}= \mathbf{W}^{\tau}_{\rm out}\cdot\mathbb{R}. 
\end{align}
and substituting the expression for the optimized output weight vector 
\begin{align} \nonumber
\mathbf{W}_{\rm out}^{\tau}=\mathbf{u}_{\tau}\cdot\mathbb{R}^{\intercal}\cdot(\mathbb{R}\cdot\mathbb{R}^{\intercal})^{-1},
\end{align}
which minimizes the mean squared error in Eq.~\eqref{eq:eqMFN3}, we obtain 
\begin{align} \label{eq:eq3}
	{\rm M}_F(\tau) \equiv {\rm M}_{F}[\mathbb{R},\mathbf{u}_{\tau}]=\frac{\mathbf{u}_{\tau}\cdot\mathbb{R}^{\intercal}\cdot(\mathbb{R}\cdot\mathbb{R}^{\intercal})^{-1}\cdot\mathbb{R}\cdot\mathbf{u}_{\tau}^{\intercal}}{\mathbf{u}_{\tau}\cdot\mathbf{u}_{\tau}^{\intercal}}.
\end{align}
where $u_{\tau}$ is the delayed scalar input time series and $\mathbb{R}$ is the reservoir state matrix. In cases where $\mathbb{R}\cdot\mathbb{R}^{\intercal}$ is not full rank, the term $(\mathbb{R}\cdot\mathbb{R}^{\intercal})^{-1}$ represents the Moore-Penrose pseudoinverse, which acts as the inverse on the complementary subspace while mapping the kernel subspace to zero.

\begin{figure*} [ht!]
\includegraphics[width=0.8\linewidth]{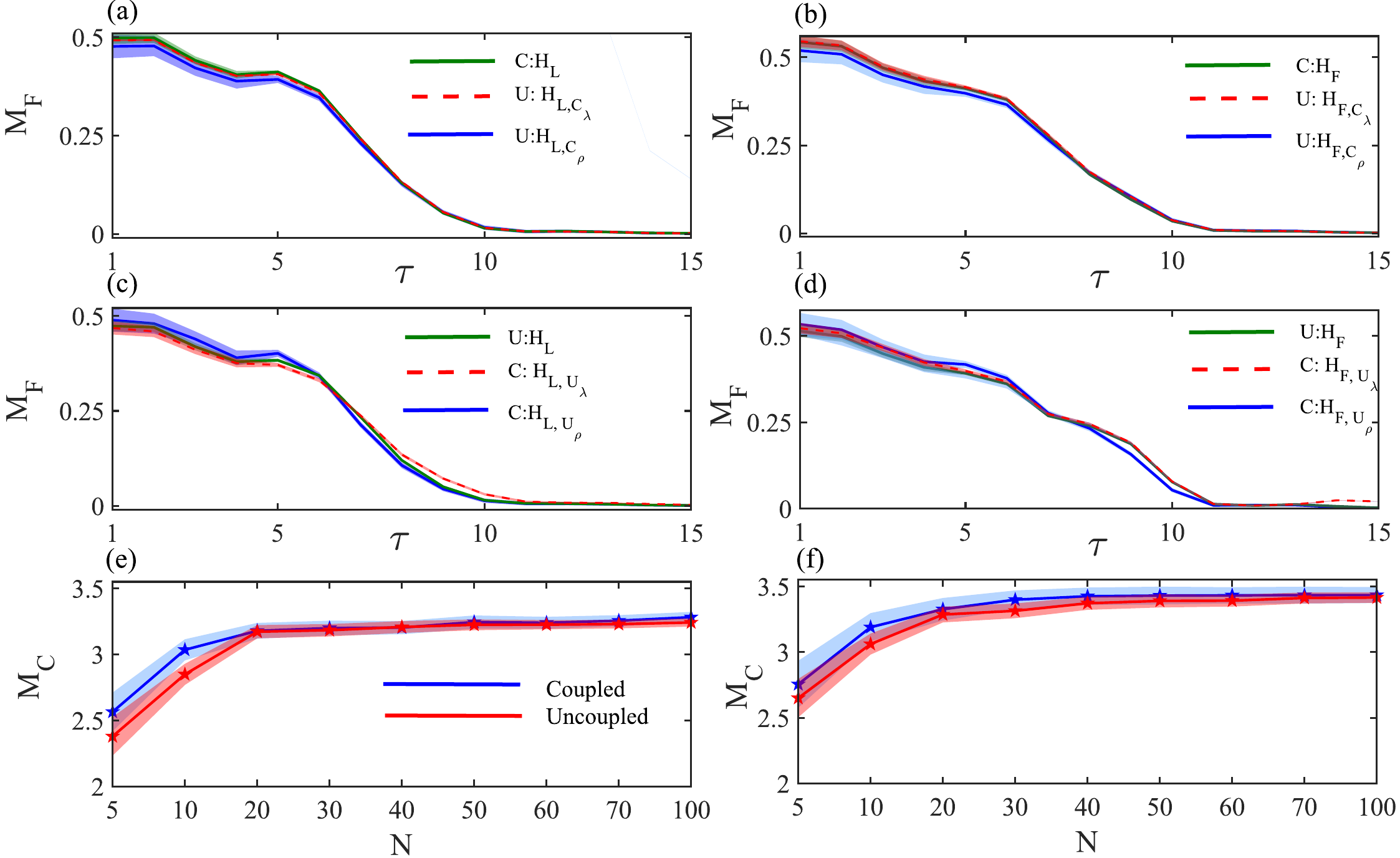}
\caption{Memory function ${\rm M}_F$ and memory capacity ${\rm M}_C$ of the vestibular reservoir computer. (a-d) Memory function ${\rm M}_F$ versus time delay for stochastic input. Panels (a, c) use hyperparameters optimized for the Lorenz system; panels (b, d) use hyperparameters optimized for the chaotic food-chain system. In panels (a) and (b), the green trace represents the baseline coupled reservoir; red points denote an uncoupled network constrained to share the same eigenvalue spectrum, while blue points denote an uncoupled network sharing only the spectral radius $\rho$ (with random eigenvalues). In panels (c) and (d), the green trace represents the baseline uncoupled reservoir, compared against coupled networks with identical eigenvalues (red) or identical spectral radius (blue). (e–f) Total memory capacity ${\rm M}_C$ as a function of reservoir size for coupled (blue) and uncoupled (red) configurations, using hyperparameters tuned on the Lorenz (e) and chaotic food-chain (f) systems. All results represent averages over 100 independent trials.}
\label{fig:MF_VRC}
\end{figure*}

We next investigate the extent to which the results derived from the linear reservoir computer extend to the nonlinear vestibular reservoir computer by computing the memory function ${\rm M}_F(\tau)$ and memory capacity ${\rm M}_C$ across different hyperparameter configurations and reservoir topologies (coupled and uncoupled). The input consists of a scalar stochastic signal, and the optimized hyperparameter values are adopted from the Lorenz and chaotic food-chain system experiments (as used in Fig.~\ref{fig:fig2}).
Now, consider a linear echo state network where the state updates according to the rule: 
\begin{align} \label{eq:RC_Linear_Update_Rule} 
\mathbf{r}(t+1)=\mathbb{A}\cdot \mathbf{r}(t) + \mathbb{W}_{\rm in}\cdot {u}(t). 
\end{align} 
To eliminate the impact of input correlations on memory capacity, we assume a purely random input signal $u(t)$. It can be shown that the function ${\rm M}_F(\tau)$ is given by (see SI Sec.~\ref{SI:Sec5}): 
\begin{align} \label{eq:eqMF1}
	{\rm M}_{F}[\mathbb{R}, \mathbf{u}_{\tau}]= \mathbf{H}_{\tau}^{\intercal}\cdot(\mathbb{H}\cdot\mathbb{H}^{\intercal})^{-1}\cdot \mathbf{H}_{\tau}
\end{align} 
yielding the following formula for ${\rm M}_{C}$: 
\begin{align}
	{\rm M}_{C}= \sum_{\tau = 0}^{T-1} {\rm M}_{F}[\mathbb{R},\mathbf{u}_{\tau}]= \mathrm{tr} \left[ \mathbb{H}^{\intercal} \cdot(\mathbb{H}\cdot\mathbb{H}^{\intercal})^{-1}\cdot \mathbb{H} \right],
\end{align} 
where $\mathbb{H}$ is the matrix of eigenvalues with dimension $N \times T$: 
\begin{align} 
	\mathbf{H}_{\tau} = \begin{bmatrix} \lambda_{1}^{\tau}, \ldots, \lambda_{N}^{\tau} \end{bmatrix}^{\intercal} 
\end{align} 
and 
\begin{align}
\mathbb{H} = \begin{bmatrix}
\lambda_{1}^{T-1} & \lambda_{1}^{T-2} & \cdots & \lambda_{1} & 1 \\
\vdots & \vdots & \ddots & \vdots & \vdots \\
\lambda_{N}^{T-1} & \lambda_{N}^{T-2} & \cdots & \lambda_{N} & 1
\end{bmatrix}.
\end{align}

Figure~\ref{fig:MC_val} displays the memory function ${\rm M}_F(\tau)$ derived theoretically [Eq.~\eqref{eq:eqMF1}] alongside simulation results [Eq.~\eqref{eq:MF_updated}] for both coupled and uncoupled linear reservoirs with the same spectral radius $\rho$. It is evident that the theoretical and numerical curves agree. Crucially, the expression for ${\rm M}_F(\tau)$ in Eq.~\eqref{eq:eqMF1} reveals that the memory function of a linear reservoir depends solely on the eigenvalues of the internal weight matrix $\mathbb{A}$. This implies that the information processing capacities of linear coupled and uncoupled reservoirs are equivalent, provided that the eigenvalues of their respective $\mathbb{A}$ matrices are identical. Figure~\ref{fig:MC_val}(c) provides further evidence for this finding, showing that the numerically obtained ${\rm M}_F$ for an uncoupled linear reservoir matches that of its coupled counterpart when their eigenvalues are identical.

\subsubsection{Memory capacity of the vestibular reservoir computer}

Figure~\ref{fig:MF_VRC} displays the memory function ${\rm M}_F(\tau)$ for coupled and uncoupled reservoirs. Panels (a,c) and (b,d) utilize hyperparameter values trained on the Lorenz and chaotic food-chain systems, respectively. Specifically, in Fig.~\ref{fig:MF_VRC}(a), the baseline model is the coupled reservoir computer, with its memory function ${\rm M}_F(\tau)$ shown in green. We compare this baseline against two distinct uncoupled configurations: one with an eigenvalue spectrum of the connectivity matrix ($\mathbb{A}$) identical to the coupled reservoir (red), and another with the same spectral radius $\rho$ as the coupled reservoir but with eigenvalues of the reservoir connectivity matrix sampled randomly (blue). The three curves exhibit reasonable agreement, indicating similar predictive capabilities across the three network configurations. Conversely, in Fig.~\ref{fig:MF_VRC}(c), the baseline model is an uncoupled reservoir (green trace). The memory functions of two coupled reservoir models, one with identical eigenvalues (red) and one with the same spectral radius (blue), are compared against it. Again, the three memory functions align closely. Figures~\ref{fig:MF_VRC}(b) and \ref{fig:MF_VRC}(d) follow the same legend scheme as (a) and (c), respectively, but utilize hyperparameters tuned on the chaotic food-chain data.

In all cases, the function ${\rm M}_F(\tau)$ is globally decreasing with respect to the time delay $\tau$, consistent with intuition. A critical observation is that the memory function of a coupled (or uncoupled) reservoir closely matches that of its uncoupled (or coupled) counterpart when their eigenvalue spectra are identical. However, when the networks share only the same spectral radius (but possess different eigenvalues), a noticeable mismatch in the memory functions appears. The overall profiles of the memory function remain consistent regardless of the specific hyperparameter values used. These findings align with the theoretical results obtained for linear reservoirs (e.g., Fig.~\ref{fig:MC_val}(c)).

Finally, we calculate the total memory capacity ${\rm M}_C$ [Eq.~\eqref{eq:MC}] as a function of network size. The results are presented in Figs.~\ref{fig:MF_VRC}(e) and \ref{fig:MF_VRC}(f) for the Lorenz and food-chain system hyperparameters, respectively. For small networks (fewer than $\approx 20$ nodes), the coupled vestibular reservoir computer exhibits a slightly higher memory capacity than its uncoupled counterpart; however, this difference diminishes as network size increases. Remarkably, the relationship between memory capacity and network size mirrors the trend observed in the prediction results (Figs.~\ref{fig:error} and \ref{fig:error1}), suggesting that memory capacity is the primary mechanism determining predictive performance. When coupled and uncoupled reservoir computers possess the same eigenvalue spectrum, their memory capacities are approximately equal. (Additional results are presented in Figs.~\ref{fig:figsi2}-\ref{fig:figsi3} and Tables~\ref{tab:tabS41}-\ref{tab:tabS42} in the Supplementary Information). Consequently, their capabilities for predicting chaotic systems are similar, as demonstrated in Figs.~\ref{fig:error} and \ref{fig:error1}. Given that physical implementations of fully coupled reservoir networks are generally significantly more challenging than uncoupled ones, our results indicate that a carefully designed uncoupled network can achieve comparable performance, making it a highly promising architecture for physical reservoir computing.

\section{Discussion} \label{sec:discussion}

This study makes two primary contributions. First, we introduce a bio-inspired physical reservoir computing model based on the intricate dynamics of the biological vestibular system, a structure recognized for maintaining stability and control in biological organisms. We term this biophysics-inspired machine learning model \textit{vestibular reservoir computing}. Unlike traditional reservoir computing frameworks primarily implemented using artificial neural networks, vestibular reservoir computing is driven by the biomechanical and neurophysiological interactions within the vestibular system. Our model captures system-level interactions between biomechanical components (semicircular canals and otolith organs) and neurophysiological processes (hair cell signal processing) while intentionally avoiding highly specific physiological details. Nonlinear reservoir dynamics emerge naturally from the interaction between external stimuli, endolymph movement, and neuronal membrane voltage changes, where the reservoir state is defined by the membrane voltage of the modeled neurons. This framework offers a perspective on how biological systems can serve as efficient, self-organizing reservoir computers for predicting nonlinear dynamical systems.

To establish a direct comparison with conventional frameworks, we first examined the performance of a coupled vestibular reservoir computer, where the reservoir consists of interconnected components with a random coupling topology. While this structure resembles traditional software-based reservoirs, physically implementing such complex interactions is highly impractical. The difficulty of engineering precise interconnections between physical components motivated our transition to an uncoupled reservoir computing framework, where each node operates independently while maintaining high computational power. Our results reveal that the uncoupled framework not only circumvents the implementation challenges of coupled reservoirs but also delivers unexpectedly robust performance. Remarkably, by employing independent reservoir nodes in parallel, we achieved accurate one-step open-loop predictions and stable long-term closed-loop forecasting, demonstrating that physical coupling is not a strict requirement for high-performance vestibular reservoir computing.

To further investigate stability and scalability, we conducted a systematic multi-step, closed-loop performance analysis by varying the number of reservoir nodes and computing statistics for short- and long-term predictions. Each configuration was tested over a large number of realizations to assess robustness. The results demonstrate that increasing the number of nodes consistently improves performance, as larger reservoirs provide richer nonlinear transformations and increased memory capacity (MC). However, the performance of uncoupled and coupled reservoirs remains comparable across different network sizes, emphasizing the efficacy of the uncoupled vestibular reservoir computer for chaotic time-series prediction. Our observations indicate that a moderate reduction in reservoir size preserves reasonable accuracy, suggesting that physical reservoir computing implementations can be optimized to use fewer computational and hardware resources without significantly compromising performance. This trade-off is particularly critical for real-world applications where minimizing hardware complexity is a key design constraint. We note that our results regarding uncoupled reservoirs with reduced sizes represent an advancement over recent works that successfully forecast chaotic time series using smaller reservoirs~\cite{Jaurigue2024,ma2023efficient}.

The second contribution of this work is the counterintuitive finding that the simplest reservoir topology, the uncoupled network, can, if properly designed, achieve prediction performance comparable to coupled topologies. This finding is of special relevance for experimental realization, as uncoupled networks are significantly easier to implement. This empirical result is supported by a theoretical analysis of memory capacity in linear reservoir computers, establishing that when an uncoupled network shares the same eigenvalues of the adjacency matrix as its coupled counterpart, it possesses a similar memory capacity. While such an analytical derivation is not feasible for the nonlinear vestibular reservoir computer, numerical evidence supports this conclusion: if the uncoupled network and its coupled counterpart possess the same eigenvalue spectrum, they exhibit approximately the same memory capacity. Since memory capacity is central to reservoir computing, this implies that specific types of uncoupled networks can yield prediction performance equivalent to coupled networks. A corollary result is that if an uncoupled and a coupled network share only the same spectral radius but have distinct eigenvalues, their memory capacities differ, typically leading to lower performance for the uncoupled reservoir. Given that the existing literature often emphasizes the role of the spectral radius, our findings highlight that the full eigenvalue spectrum is equally influential.

While our computations focus on the vestibular framework, these results support the broader conjecture that within physical reservoir computing, simply matching the spectral radius does not guarantee performance; however, aligning the entire eigenvalue spectrum enables uncoupled reservoirs to attain performance levels comparable to coupled systems. This concept of exploiting uncoupled networks can be instrumental for other reservoir types and supports the physical implementation of reservoirs with independent nodes.

Collectively, these findings serve as a proof of concept that the biological vestibular system can function as a physical reservoir computer, offering a biologically plausible mechanism for real-time dynamical processing. By demonstrating high performance without complex interconnectivity, this work opens new avenues for bio-inspired computing architectures, potentially leading to neuromorphic implementations that leverage natural physical dynamics for efficient computation. Our results are among the first in the context of uncoupled reservoir computing to highlight its applicability beyond basic tasks. Future work may modify the bio-inspired physical reservoir computer by incorporating spiking behavior to increase biological realism. Currently, our model is set to produce a fixed-point steady state; transitioning to a spiking regime will introduce new challenges, particularly in maintaining the echo state property, which warrants further study.

\section{Methods} \label{sec:methods}

\subsection{Mathematical Models for Chaotic Time Series}\label{M_model}

We focus on chaotic time series as input data to evaluate the performance of the proposed vestibular reservoir computer framework. Specifically, we employ two different chaotic systems, the Lorenz system and the chaotic food-chain system, as sources of chaotic signals, representing benchmarks for complex nonlinear time-series prediction.

The Lorenz system is defined by~\cite{Lorenz:1963}: 
\begin{subequations}\label{eq:eqinput1} 
\begin{align} 
\dot{u}_1 = \sigma (u_2 - u_1), \ \ \dot{u}_2 = u_1(\rho - u_3) - u_2, \ \ \dot{u}_3 = u_1 u_2 - \beta u_3, 
\end{align} 
\end{subequations} 
which is simulated with a time step of $h = 10^{-3}$. The system parameters are set to $\sigma=10$, $\rho=28$, and $\beta=8/3$. The three state variables $(u_1, u_2, u_3)$ are sampled at intervals of $\Delta t = 0.1$ and normalized to the range $[0,1]$.

The dynamics of the chaotic food-chain system are governed by~\cite{HH:1991,MY:1994}:
\begin{subequations}    \label{eq:eqinput2}
        \begin{align}
        \dot{u}_1 &= u_1 \left( 1 - \frac{u_1}{K} \right) - \frac{a_1 b_1 \, u_2 u_1}{u_1 + a_{10}}, \\
        \dot{u}_2 &= a_1 u_2 \left( \frac{b_1 u_1}{u_1 + a_{10}} - 1 \right) - \frac{a_2 b_2 \, u_3 u_2}{u_2 + a_{20}}, \\
        \dot{u}_3 &= a_2 u_3 \left( \frac{b_2 u_2}{u_2 + a_{20}} - 1 \right).
        \end{align}
\end{subequations}
where $u_1$, $u_2$, and $u_3$ denote the population densities of the resource, consumer, and predator species, respectively. The system parameters are set to~\cite{MY:1994}: $K=0.98$, $a_{1}=0.4$, $b_{1}=2.009$, $a_{2}=0.08$, $b_{2}=2.876$, $a_{10}=0.16129$, and $b_{20}=0.5$.

In our approach, we evaluate the performance of the vestibular reservoir computer for both coupled and uncoupled configurations by adjusting the structure of the reservoir matrix $\mathbb{A}$ in Eq.~\eqref{SCCFHN} separately for the Lorenz and chaotic food-chain inputs. The uncoupled configurations replace complex interconnected networks with a collection of independent dynamical elements. The input signals can be processed either in parallel, using multiple independent reservoirs simultaneously, or in a sequential manner, where multiple independent experiments are conducted with a single reservoir and the results are subsequently aggregated. An experiment consists of training, validation, and testing phases. The training and validation phases constitute the open-loop process, while the testing phase constitutes the closed-loop process, as detailed below.

\subsection{Reservoir Computing} \label{M_RC}

A reservoir computer comprises an input layer, a hidden layer containing a recurrent neural network, and an output layer. At the conclusion of the training phase, the output weights, specifically, the elements of the output matrix $\mathbb{W}_{\rm out}$, are determined via linear regression to subsequently generate short-term and long-term predictions.

The entire training and prediction process using vestibular reservoir computing can be divided into an open-loop phase and a closed-loop phase. In the open-loop phase, the reservoir is driven by an external input sequence, and its responses are recorded for training and validation. The input is mapped into the reservoir via an input weight matrix $\mathbb{W}{\text{in}} \in \mathcal{R}^{N \times D}$, where $N$ is the reservoir size and the input signal is three-dimensional ($D = 3$) for the two chaotic target systems. The external input $\mathbf{u}=[u_{1},u_{2},u_{3}]^{\intercal}$, obtained by solving the Lorenz system [Eq.~\eqref{eq:eqinput1}] or the chaotic food-chain model [Eq.~\eqref{eq:eqinput2}], is a sequence of length $L_{\text{transient}} + L_{\text{train}} + L_{\text{validation}}$. This sequence is fed into the reservoir described by:
\begin{align} \nonumber 
\dot{\mathbf{r}} = \tau \mathbf{f}(\mathbf{r},\mathbf{u}), 
\end{align} 
where $\tau$ denotes the time constant, $\mathbf{r} \in \mathcal{R}^{N}$ represents the reservoir state, and $\mathbf{f}$ is the activation function obtained by solving Eq.~\eqref{SCCFHN} for the vestibular reservoir computer. The input data are normalized, and the first $L_{\text{transient}}$ responses are discarded to ensure the reservoir state is independent of initial conditions. We use $L_{transient}=10000$ for both the Lorenz and chaotic food-chain systems. The remaining $L_{\text{train}} + L_{\text{validation}}$ responses are stored in the reservoir state matrix $\mathbb{R}_{\text{int}} \in \mathcal{R}^{N \times (L{\text{train}} + L_{\text{validation}})}$, where the entry $r_{j,k}$ represents the $j$-th sampled response of the reservoir to the $(L_{\text{transient}} + k)$-th input. The first $L_{\text{train}}$ columns of matrix $\mathbb{R}$, denoted as $\mathbb{R}_{\text{train}}$, are utilized for training. The remaining $L_{\text{validation}}$ columns of $\mathbb{R}$ (denoted as $\mathbb{R}_{\text{validation}}$) are used for validation. The predicted validation output is given by: 
\begin{equation} \label{eq:RC_Validation_Output} 
\hat{\mathbf{y}}_{\text{validation}} = \mathbb{W}_{\text{out}}\cdot \mathbb{R}_{\text{validation}}.  
\end{equation}
We set $L_{train}=10000$ and $50000$ for the Lorenz and chaotic food-chain systems, respectively, whereas $L_{validation}=5000$ for both target systems. To enhance reservoir performance and improve the condition number of $\mathbb{R}$, we augment the state matrix as $\mathbb{R} = [\mathbb {R}_{\text{int}}; \mathbb{R}_s]$, where $\mathbb{R}_s$ is a matrix of the same size as $\mathbb{R}_{\text{int}}$ containing the squared values of the entries in $\mathbb{R}_{\text{int}}$. This augmentation helps mitigate issues such as ``mirror attractors'' and enhances predictive accuracy. Consequently, the output matrix is augmented to have dimensions $D\times (2N)$.

According to reservoir computing principles, the output is a linear combination of reservoir states. The optimal output weight matrix $\mathbb{W}{\text{out}}$ is obtained using ridge regression: 
\begin{equation} \label{eq:RC_Ridge_Regression}  
\mathbb{W}_{\text{out}} = \mathbb{Y}\cdot\mathbb{R}^T \cdot(\mathbb{R}\cdot\mathbb{R}^T + \lambda \mathbb{I})^{-1} ,  
\end{equation}
where $\lambda$ is the Tikhonov regularization parameter, $\mathbb{Y}$ is the matrix containing target output vectors, and $I$ is the identity matrix. The predicted output for training is then given by: 
\begin{align} \label{eq:RC_Training_Output}
\hat{\mathbf{y}}_{\text{train}} = \mathbb{W}_{\text{out}}\cdot\mathbb{R}_{\text{train}}.
\end{align}
Training and validation errors are computed for the trained reservoir computer, and these errors are minimized via hyperparameter optimization to ensure generalization before proceeding to the closed-loop phase. Once the output weight matrix $\mathbb{W}{\text{out}}$ is determined, the system transitions to the closed-loop phase, where the reservoir operates autonomously. Instead of using an external input $\mathbf{u}$, the predicted reservoir output is fed back as the input for the next time step. This self-sustained prediction mechanism enables long-term autonomous time-series forecasting. The closed-loop dynamics are governed by: 
\begin{align}
\mathbf{u}_{\text{C}}(t+1) &= \mathbb{W}_{\text{out}} \cdot \big( \mathbf{r}(t), \mathbf{r}^2(t) \big),\\
\dot{\mathbf{r}}(t) &= \tau \mathbf{f}\big(\mathbf{r}(t), \mathbf{u}_{\text{C}}(t)\big),
\end{align}
where $\mathbf{r}^2(t)$ denotes the element-wise squared vector of $\mathbf{r}(t)$, consistent with the augmented state matrix $\mathbb{R}$ used during training. The objective of this phase is for the autonomous system to emulate the dynamics of the original chaotic system that generated the training data.

Long-term prediction accuracy is assessed by examining whether the reservoir-predicted attractor remains in the vicinity of the ground-truth attractor. Statistical measures such as the largest Lyapunov exponent, the DV value, and the KL divergence are employed for this purpose. Further methodological details, including hyperparameters of the vestibular reservoir computer and predictive statistics, are presented in the SI (Secs.~\ref{secH}-\ref{si_statistic}).


\section*{Acknowledgements}

This material is based upon work supported by the Laboratory University Collaboration Initiative (LUCI) program, through an award made by the Office of the Under Secretary of War for Research and Engineering \big(OUSW(R\&E)\big), Science and Technology (S\&T)/Foundations. This work was also supported by the US Army Research Office under Grants No.~W911NF-24-2-0228 and No.~W911NF-26-2-A002.

\section*{Author contributions}

S.P., S.D., M.H. and Y.-C.L. designed the research project, the models, and methods. S.P. and S.D. and performed the computations. S.P., S.D., M.H. and Y.-C.L. analyzed the data. S.P., S.D., and Y.-C.L. wrote the paper. Y.-C.L. edited the manuscript.

\section*{Declarations}

The authors declare no competing interests.

\section*{Additional information}

Supplementary information available at~Supplementary Information.

\section*{Data availability} 

Codes and data are available in a Github repository (\url{https://github.com/SMITA1996/Vestibular-reservoir-computing}).

\section*{Correspondence}

Correspondence and requests for materials should be addressed to Ying-Cheng.Lai@asu.edu.

\bibliography{VRC}

@article{steinhausen1933,
  title={Ueber die Beobachtung der Cupula in den Bogengangsampullen des Labyrinths des lebenden Hechts.},
  author={Steinhausen, Wilhelm},
  journal={Pfl{\"u}gers Archiv f{\"u}r die gesamte Physiologie des Menschen und der Tiere},
  year={1933}
}

@article{vanEgmond1949,
  title = {The mechanics of the semicircular canal},
  volume = {110},
  ISSN = {1469-7793},
  url = {http://dx.doi.org/10.1113/jphysiol.1949.sp004416},
  DOI = {10.1113/jphysiol.1949.sp004416},
  number = {1–2},
  journal = {The Journal of Physiology},
  publisher = {Wiley},
  author = {van Egmond,  A. A. J. and Groen,  J. J. and Jongkees,  L. B. W.},
  year = {1949},
  month = dec,
  pages = {1–17}
}

@article{groen1956,
  title={The semicircular canal system of the organs of equilibrium-I},
  author={Groen, JJ},
  journal={Physics in Medicine \& Biology},
  volume={1},
  number={2},
  pages={103},
  year={1956},
  publisher={IOP Publishing}
}

@article{groen1957,
  title={The semicircular canal system of the organs of equilibrium-II},
  author={Groen, JJ},
  journal={Physics in Medicine \& Biology},
  volume={1},
  number={3},
  pages={225},
  year={1957},
  publisher={IOP Publishing}
}

@article{young1969,
  title={The current status of vestibular system models},
  author={Young, Laurence R},
  journal={Automatica},
  volume={5},
  number={3},
  pages={369--383},
  year={1969},
  publisher={Elsevier}
}

@article{schmid1971,
  title={Mathematical modelling: a contribution to clinical vestibular analysis},
  author={Schmid, R and Stefanelli, M and Mira, E},
  journal={Acta oto-laryngologica},
  volume={72},
  number={1-6},
  pages={292--302},
  year={1971},
  publisher={Taylor \& Francis}
}

@article{benson1974,
  title={A systems concept of the vestibular organs},
  author={Benson, AJ and Bischof, N and Collins, WE and Fregly, AR and Graybiel, A and Guedry, FE and Johnson, WH and Jongkees, LBW and Kornhuber, HH and Mayne, R and others},
  journal={Vestibular system part 2: psychophysics, applied aspects and general interpretations},
  pages={493--580},
  year={1974},
  publisher={Springer}
}

@article{gernandt1949,
  title={Response of mammalian vestibular neurons to horizontal rotation and caloric stimulation},
  author={Gernandt, Bo},
  journal={Journal of neurophysiology},
  volume={12},
  number={3},
  pages={173--184},
  year={1949}
}

@incollection{rabbitt2004,
  title={Biomechanics of the semicircular canals and otolith organs},
  author={Rabbitt, Richard D and Damiano, Edward R and Grant, J Wallace},
  booktitle={The vestibular system},
  pages={153--201},
  year={2004},
  publisher={Springer}
}

@phdthesis{meiry1965,
  title={The vestibular system and human dynamic space orientation.},
  author={Meiry, Jacob Leon},
  year={1965},
  school={Massachusetts Institute of Technology}
}

@techreport{houck2005,
  title={Motion cueing algorithm development: Human-centered linear and nonlinear approaches},
  author={Houck, Jacob A and Telban, Robert J and Cardullo, Frank M},
  journal={},
  volume={},
  pages={},
  institution={NASA},
  url={https://api.semanticscholar.org/CorpusID:59776455},
  year={2005}
}

@article{sadovnichy2007,
  title={A mathematical model of the response of the semicircular canal and otolith to vestibular system rotation under gravity},
  author={Sadovnichy, VA and Alexandrov, VV and Soto, E and Alexandrova, TB and Astakhova, TG and Vega, R and Kulikovskaya, NV and Kurilov, VI and Migunov, SS and Shulenina, NE},
  journal={Journal of Mathematical Sciences},
  volume={146},
  pages={5938--5947},
  year={2007},
  publisher={Springer}
}

@article{gastaldi2009,
  title={Vestibular apparatus: dynamic model of the semicircular canals},
  author={Gastaldi, Laura and Pastorelli, S and Sorli, Massimo and others},
  journal={WIT Trans. Biomed. Heal},
  volume={13},
  pages={223--234},
  year={2009}
}

@techreport{peters1969,
  title={Dynamics of the vestibular system and their relation to motion perception, spatial disorientation, and illusions},
  author={Peters, Richard A},
  year={1969},
  institution={NASA}
}

@article{stewart2009,
  title={Python scripting in the Nengo simulator},
  author={Stewart, Terrence C and Tripp, Bryan and Eliasmith, Chris},
  journal={Frontiers in neuroinformatics},
  volume={3},
  pages={359},
  year={2009},
  publisher={Frontiers}
}

@article{bradshaw2010,
  title={A mathematical model of human semicircular canal geometry: a new basis for interpreting vestibular physiology},
  author={Bradshaw, Andrew P and Curthoys, Ian S and Todd, Michael J and Magnussen, John S and Taubman, David S and Aw, Swee T and Halmagyi, G Michael},
  journal={Journal of the Association for Research in Otolaryngology},
  volume={11},
  pages={145--159},
  year={2010},
  publisher={Springer}
}

@article{mccollum2010,
  title={Symmetries of the central vestibular system: forming movements for gravity and a three-dimensional world},
  author={McCollum, Gin and Hanes, Douglas A},
  journal={Symmetry},
  volume={2},
  number={3},
  pages={1544--1558},
  year={2010},
  publisher={Molecular Diversity Preservation International}
}

@article{asadi2017,
  title={Semicircular canal modeling in human perception},
  author={Asadi, Houshyar and Mohamed, Shady and Lim, Chee Peng and Nahavandi, Saeid and Nalivaiko, Eugene},
  journal={Reviews in the Neurosciences},
  volume={28},
  number={5},
  pages={537--549},
  year={2017},
  publisher={De Gruyter}
}

@inproceedings{Momani2018,
  title = {A Review of the Recent Literature on the Mathematical Modeling of the Vestibular System},
  url = {http://dx.doi.org/10.2514/6.2018-0114},
  DOI = {10.2514/6.2018-0114},
  booktitle = {2018 AIAA Modeling and Simulation Technologies Conference},
  publisher = {American Institute of Aeronautics and Astronautics},
  author = {Momani,  Ahmad Q. and Cardullo,  Frank M.},
  year = {2018},
  month = jan 
}

@article{paulin2019,
  title={Models of vestibular semicircular canal afferent neuron firing activity},
  author={Paulin, Michael G and Hoffman, Larry F},
  journal={Journal of Neurophysiology},
  volume={122},
  number={6},
  pages={2548--2567},
  year={2019},
  publisher={American Physiological Society Bethesda, MD}
}

@article{ramat2019,
  title={Understanding the rotational vestibular ocular reflex: from differential equations to laplace transforms},
  author={Ramat, Stefano},
  journal={Progress in Brain Research},
  volume={248},
  pages={29--44},
  year={2019},
  publisher={Elsevier}
}

@article{lu2021,
  title={Modeling the Human Vestibular System as a Controlled System},
  author={Lu, Fangyi and Nguyen, Mary and Nguyen, Yen and Karthik, R and Tcheng, Zoe},
  journal={Department of Bioengineering-University of California, San Diego, San Diego},
  year={2021}
}

@article{minyailo2024,
  title={Linear Mathematical Model of Relation between Extraocular Muscles and Vestibular System},
  author={Minyailo, Ya Yu and Kruchinina, AP},
  journal={Moscow University Mechanics Bulletin},
  volume={79},
  number={1},
  pages={12--20},
  year={2024},
  publisher={Springer}
}

@article{asadnia2016,
  title={From biological cilia to artificial flow sensors: Biomimetic soft polymer nanosensors with high sensing performance},
  author={Asadnia, Mohsen and Kottapalli, Ajay Giri Prakash and Karavitaki, K Domenica and Warkiani, Majid Ebrahimi and Miao, Jianmin and Corey, David P and Triantafyllou, Michael},
  journal={Scientific reports},
  volume={6},
  number={1},
  pages={1--13},
  year={2016},
  publisher={Nature Publishing Group}
}

@book{goldberg2012,
  title={The vestibular system: a sixth sense},
  author={Goldberg, Jay M},
  year={2012},
  publisher={Oxford University Press, USA}
}

@article{khan2013,
  title={Anatomy of the vestibular system: a review},
  author={Khan, Sarah and Chang, Richard},
  journal={NeuroRehabilitation},
  volume={32},
  number={3},
  pages={437--443},
  year={2013},
  publisher={IOS Press}
}

@book{highstein2004,
  title={The vestibular system},
  author={Highstein, Stephen M and Fay, Richard R and Popper, Arthur N},
  volume={24},
  year={2004},
  publisher={Springer}
}

@article{cullen2019,
  title={Vestibular processing during natural self-motion: implications for perception and action},
  author={Cullen, Kathleen E},
  journal={Nature Reviews Neuroscience},
  volume={20},
  number={6},
  pages={346--363},
  year={2019},
  publisher={Nature Publishing Group UK London}
}

@article{guth1998,
  title={The vestibular hair cells: post-transductional signal processing},
  author={Guth, PS and Perin, Paola and Norris, CH and Valli, P},
  journal={Progress in neurobiology},
  volume={54},
  number={2},
  pages={193--247},
  year={1998},
  publisher={Elsevier}
}

@article{hudspeth1979,
  title={Stereocilia mediate transduction in vertebrate hair cells (auditory system/cilium/vestibular system).},
  author={Hudspeth, AJ and Jacobs, R},
  journal={Proceedings of the National Academy of Sciences},
  volume={76},
  number={3},
  pages={1506--1509},
  year={1979},
  publisher={National Acad Sciences}
}

@article{wall2003,
  title={Vestibular prostheses: the engineering and biomedical issues},
  author={Wall Iii, C and Merfeld, DM and Rauch, SD and Black, FO},
  journal={Journal of Vestibular Research},
  volume={12},
  number={2-3},
  pages={95--113},
  year={2003},
  publisher={IOS Press}
}

@article{moshizi2022,
  title={Recent advancements in bioelectronic devices to interface with the peripheral vestibular system},
  author={Moshizi, Sajad Abolpour and Pastras, Christopher John and Sharma, Rajni and Mahmud, MA Parvez and Ryan, Rachel and Razmjou, Amir and Asadnia, Mohsen},
  journal={Biosensors and Bioelectronics},
  volume={214},
  pages={114521},
  year={2022},
  publisher={Elsevier}
}

@article{lacour1993,
  title={Vestibular control of posture and gait.},
  author={Lacour, M and Borel, L},
  journal={Archives Italiennes de Biologie},
  volume={131},
  number={2},
  pages={81--104},
  year={1993}
}

@article{whitney2009,
  title={Gaze stabilization and gait performance in vestibular dysfunction},
  author={Whitney, Susan L and Marchetti, Gregory F and Pritcher, Miranda and Furman, Joseph M},
  journal={Gait \& posture},
  volume={29},
  number={2},
  pages={194--198},
  year={2009},
  publisher={Elsevier}
}

@inproceedings{patane2004,
  title={Design and development of a biologically-inspired artificial vestibular system for robot heads},
  author={Patane, Francesco and Laschi, Cecilia and Miwa, Hiroyasu and Guglielmelli, Eugenio and Dario, Paolo and Takanishi, Atsuo},
  booktitle={2004 IEEE/RSJ International Conference on Intelligent Robots and Systems (IROS)(IEEE Cat. No. 04CH37566)},
  volume={2},
  pages={1317--1322},
  year={2004},
  organization={IEEE}
}

@article{mergner2009,
  title = {Vestibular humanoid postural control},
  volume = {103},
  ISSN = {0928-4257},
  number = {3–5},
  journal = {Journal of Physiology-Paris},
  publisher = {Elsevier BV},
  author = {Mergner, Thomas and Schweigart, Georg and Fennell, Luminous},
  year = {2009},
  pages = {178–194}
}

@article{tin2005,
  title={Internal models in sensorimotor integration: perspectives from adaptive control theory},
  author={Tin, Chung and Poon, Chi-Sang},
  journal={Journal of Neural Engineering},
  volume={2},
  number={3},
  pages={S147},
  year={2005},
  publisher={IOP Publishing}
}

@article{fransson2003,
  title={Postural control adaptation during galvanic vestibular and vibratory proprioceptive stimulation},
  author={Fransson, P-A and Hafstrom, Anna and Karlberg, Mikael and Magnusson, M{\aa}ns and Tjader, Annika and Johansson, Rolf},
  journal={IEEE Transactions on Biomedical Engineering},
  volume={50},
  number={12},
  pages={1310--1319},
  year={2003},
  publisher={IEEE}
}

@article{whitney2016,
  title={Recent evidence about the effectiveness of vestibular rehabilitation},
  author={Whitney, Susan L and Alghadir, Ahmad H and Anwer, Shahnawaz},
  journal={Current treatment options in neurology},
  volume={18},
  pages={1--15},
  year={2016},
  publisher={Springer}
}

@article{sulway2019,
  title={Advances in vestibular rehabilitation},
  author={Sulway, Shaleen and Whitney, Susan L},
  journal={Vestibular Disorders},
  volume={82},
  pages={164--169},
  year={2019},
  publisher={Karger Publishers}
}

@article{alexandrov2007,
  title={Information process in vestibular system},
  author={Alexandrov, VV and Alexandrova, TB and Vega, R and Castillo-Quiroz, G and Angeles-Vazquez, A and Reyes-Romero, M and Soto, E},
  journal={WSEAS Transactions in Biology and Medicine},
  volume={12},
  number={4},
  pages={193--203},
  year={2007},
  publisher={Citeseer}
}

@article{moore2004,
  title={The linear and non-linear relationships between action potential discharge rates and membrane potential in model vestibular neurons.},
  author={Moore, LE and Hachemaoui, M and Idoux, E and Vibert, N and Vidal, PP},
  journal={Nonlinear Studies},
  volume={11},
  number={3},
  year={2004}
}

@article{clark2019,
  title={Mathematical models for dynamic, multisensory spatial orientation perception},
  author={Clark, Torin K and Newman, Michael C and Karmali, Faisal and Oman, Charles M and Merfeld, Daniel M},
  journal={Progress in brain research},
  volume={248},
  pages={65--90},
  year={2019},
  publisher={Elsevier}
}

@article{glasauer2018,
  title={Neuronal network-based mathematical modeling of perceived verticality in acute unilateral vestibular lesions: from nerve to thalamus and cortex},
  author={Glasauer, Stefan and Dieterich, Marianne and Brandt, Thomas},
  journal={Journal of neurology},
  volume={265},
  number={Suppl 1},
  pages={101--112},
  year={2018},
  publisher={Springer}
}

@article{green2010,
  title={Internal models and neural computation in the vestibular system},
  author={Green, Andrea M and Angelaki, Dora E},
  journal={Experimental brain research},
  volume={200},
  pages={197--222},
  year={2010},
  publisher={Springer}
}

@inproceedings{aleksandrov2014,
  title={Mathematical modeling of output signal for the correction of the vestibular system inertial biosensors},
  author={Aleksandrov, Vladimir V and Romero, Maribel Reyes and Soto, Enrique and Vega, Rosario and Alexandrova, Tamara B and Bugrov, Dmitriy I and Lebedev, Anton V and Lemak, Stepan S and Tikhonova, Katerina V},
  booktitle={2014 International Symposium on Inertial Sensors and Systems (INERTIAL)},
  pages={1--4},
  year={2014},
  organization={IEEE}
}

@article{santos2017,
  title={Biomechanical study of the vestibular system of the inner ear using a numerical method},
  author={Santos, Carla F and Belinha, Jorge and Gentil, Fernanda and Parente, Marco and Areias, Bruno and Jorge, Renato Natal},
  journal={Procedia IUTAM},
  volume={24},
  pages={30--37},
  year={2017},
  publisher={Elsevier}
}

@article{canelo2018,
  title={Modeling of the human vestibular system and integration in a simulator for the study of orientation and balance control},
  author={Canelo, {\'A}ngel and Tejado Balsera, In{\'e}s and Traver Becerra, Jos{\'e} Emilio and Vinagre Jara, Blas Manuel and Nuevo Gallardo, Cristina},
  journal={Actas de las XXXIX Jornadas de Autom{\'a}tica, Badajoz, 5-7 de Septiembre de 2018},
  year={2018},
  publisher={Universidad de Extremadura}
}

@article{straka2021,
  title={Translations of Steinhausen's publications provide insight into their contributions to peripheral vestibular neuroscience},
  author={Straka, Hans and Paulin, Michael G and Hoffman, Larry F},
  journal={Frontiers in Neurology},
  volume={12},
  pages={676723},
  year={2021},
  publisher={Frontiers Media SA}
}

@article{jaeger2004,
  title={Harnessing nonlinearity: Predicting chaotic systems and saving energy in wireless communication},
  author={Jaeger, Herbert and Haas, Harald},
  journal={science},
  volume={304},
  number={5667},
  pages={78--80},
  year={2004},
  publisher={American Association for the Advancement of Science}
}

@article{lukovsevivcius2009,
  title={Reservoir computing approaches to recurrent neural network training},
  author={Luko{\v{s}}evi{\v{c}}ius, Mantas and Jaeger, Herbert},
  journal={Computer science review},
  volume={3},
  number={3},
  pages={127--149},
  year={2009},
  publisher={Elsevier}
}

@article{MNM:2002,
title={Real-time computing without stable states: A new framework for neural computation based on perturbations},
author={Maass, Wolfgang and Natschl{\"a}ger, Thomas and Markram, Henry},
journal={Neural Comput.},
volume={14},
number={11},
pages={2531--2560},
year={2002},
publisher={MIT Press}
}

@article{MJ:2013,
  title = {Echo State Property Linked to an Input: Exploring a Fundamental Characteristic of Recurrent Neural Networks},
  author = {G. Manjunath and H. Jaeger},
  journal = {Neur. Comp.},
  volume = {25},
  pages = {671-696},
  year = {2013}
}

@article{PLHGO:2017,
  title = {Using machine learning to replicate chaotic attractors and calculate {Lyapunov} exponents from data},
  author = {Pathak, Jaideep and Lu, Zhixin and Hunt, Brian and Girvan, Michelle and Ott, Edward},
  journal = {Chaos},
  volume = {27},
  pages = {121102},
  year = {2017}
}

@article{PHGLO:2018,
  title = {Model-Free Prediction of Large Spatiotemporally Chaotic Systems from Data: A Reservoir Computing Approach},
  author = {Pathak, Jaideep and Hunt, Brian and Girvan, Michelle and Lu, Zhixin and Ott, Edward},
  journal = {Phys. Rev. Lett.},
  volume = {120},
  issue = {2},
  pages = {024102},
  numpages = {5},
  year = {2018},
  month = {Jan},
  publisher = {American Physical Society},
  doi = {10.1103/PhysRevLett.120.024102}
}

@article{JL:2019,
  title = {Model-free prediction of spatiotemporal dynamical systems with recurrent neural networks: Role of network spectral radius},
  author = {Jiang, Junjie and Lai, Ying-Cheng},
  journal = {Phys. Rev. Research},
  volume = {1},
  issue = {3},
  pages = {033056},
  numpages = {14},
  year = {2019},
  month = {Oct},
  publisher = {American Physical Society},
  doi = {10.1103/PhysRevResearch.1.033056}
}

@article{TYHNKTNNH:2019,
title={Recent advances in physical reservoir computing: A review},
author={Tanaka, Gouhei and Yamane, Toshiyuki and H{\'e}roux, Jean Benoit and Nakane, Ryosho and Kanazawa, Naoki and Takeda, Seiji and Numata, Hidetoshi and Nakano, Daiju and Hirose, Akira},
journal={Neu. Net.},
volume={115},
pages={100--123},
year={2019},
publisher={Elsevier}
}

@article{PCGPO:2021,
  title={Using machine learning to predict statistical properties of non-stationary dynamical processes: System climate, regime transitions, and the effect of stochasticity},
  author={Patel, Dhruvit and Canaday, Daniel and Girvan, Michelle and Pomerance, Andrew and Ott, Edward},
  journal={Chaos},
  volume={31},
  number={3},
  pages={033149},
  year={2021},
  publisher={AIP Publishing LLC}
}

@article{Bollt:2021,
title={On explaining the surprising success of reservoir computing forecaster of chaos? The universal machine learning dynamical system with contrast to VAR and DMD},
author={Bollt, Erik},
journal={Chaos},
volume={31},
number={1},
pages={013108},
year={2021},
publisher={AIP Publishing LLC}
}

@article{GBGB:2021,
title={Next generation reservoir computing},
author={Gauthier, Daniel J and Bollt, Erik and Griffith, Aaron and Barbosa, Wendson AS},
journal={Nat. Commun.},
volume={12},
number={1},
pages={1--8},
year={2021},
publisher={Nature Publishing Group}
}

@inproceedings{Schrauwen2007,
  author       = {Schrauwen, Benjamin and Verstraeten, David and Van Campenhout, Jan},
  booktitle    = {Proceedings of the 15th European Symposium on Artificial Neural Networks. p. 471-482 2007},
  pages        = {471--482},
  title        = {An overview of reservoir computing: theory, applications and implementations},
  url          = {http://doi.org/1854/11063},
  year         = {2007},
}

@article{DVBSMDDS:2007,
title = {An experimental unification of reservoir computing methods},
journal = {Neural Netw.},
volume = {20},
number = {3},
pages = {391-403},
year = {2007},
note = {Echo State Networks and Liquid State Machines},
issn = {0893-6080},
doi = {https://doi.org/10.1016/j.neunet.2007.04.003},
url = {https://www.sciencedirect.com/science/article/pii/S089360800700038X},
author = {D. Verstraeten and B. Schrauwen and M. D’Haene and D. Stroobandt},
}

@book{NKFI:2021,
author={Nakajima, Kohei and Fischer, Ingo},
editor="Meyers, Robert A.",
title={Reservoir computing},
year={2021},
publisher={Springer Singapore},
address="Singapore",
doi={https://doi.org/10.1007/978-981-13-1687-6},
}

@article{NK:2020,
  title={Physical reservoir computing—an introductory perspective},
  author={Nakajima, Kohei},
  journal={Jpn. J. Appl. Phys.},
  volume={59},
  number={6},
  pages={060501},
  year={2020},
  publisher={IOP Publishing}
}

@article{stepney2024,
  title={Physical reservoir computing: a tutorial},
  author={Stepney, Susan},
  journal={Natural Computing},
  pages={1--21},
  year={2024},
  publisher={Springer}
}

@article{appeltant2011,
  title={Information processing using a single dynamical node as complex system},
  author={Appeltant, Lennert and Soriano, Miguel Cornelles and Van der Sande, Guy and Danckaert, Jan and Massar, Serge and Dambre, Joni and Schrauwen, Benjamin and Mirasso, Claudio R and Fischer, Ingo},
  journal={Nature communications},
  volume={2},
  number={1},
  pages={468},
  year={2011},
  publisher={Nature Publishing Group UK London}
}

@article{moran2023,
  title={Hardware-optimized reservoir computing system for edge intelligence applications},
  author={Mor{\'a}n, Alejandro and Canals, Vincent and Galan-Prado, Fabio and Frasser, Christian F and Radhakrishnan, Dhinakar and Safavi, Saeid and Rossell{\'o}, Josep L},
  journal={Cognitive Computation},
  pages={1--9},
  year={2023},
  publisher={Springer}
}

@article{milano2022,
  title={In materia reservoir computing with a fully memristive architecture based on self-organizing nanowire networks},
  author={Milano, Gianluca and Pedretti, Giacomo and Montano, Kevin and Ricci, Saverio and Hashemkhani, Shahin and Boarino, Luca and Ielmini, Daniele and Ricciardi, Carlo},
  journal={Nature materials},
  volume={21},
  number={2},
  pages={195--202},
  year={2022},
  publisher={Nature Publishing Group UK London}
}

@article{zhong2022,
  title={A memristor-based analogue reservoir computing system for real-time and power-efficient signal processing},
  author={Zhong, Yanan and Tang, Jianshi and Li, Xinyi and Liang, Xiangpeng and Liu, Zhengwu and Li, Yijun and Xi, Yue and Yao, Peng and Hao, Zhenqi and Gao, Bin and others},
  journal={Nature Electronics},
  volume={5},
  number={10},
  pages={672--681},
  year={2022},
  publisher={Nature Publishing Group UK London}
}

@article{liang2024,
  title={Physical reservoir computing with emerging electronics},
  author={Liang, Xiangpeng and Tang, Jianshi and Zhong, Yanan and Gao, Bin and Qian, He and Wu, Huaqiang},
  journal={Nature Electronics},
  volume={7},
  number={3},
  pages={193--206},
  year={2024},
  publisher={Nature Publishing Group UK London}
}

@article{rajib2022,
  title={Skyrmion based energy-efficient straintronic physical reservoir computing},
  author={Rajib, Md Mahadi and Al Misba, Walid and Chowdhury, Md Fahim F and Alam, Muhammad Sabbir and Atulasimha, Jayasimha},
  journal={Neuromorphic Computing and Engineering},
  volume={2},
  number={4},
  pages={044011},
  year={2022},
  publisher={IOP Publishing}
}

@article{picco2024,
  title={Deep photonic reservoir computer for speech recognition},
  author={Picco, Enrico and Lupo, Alessandro and Massar, Serge},
  journal={IEEE Transactions on Neural Networks and Learning Systems},
  year={2024},
  publisher={IEEE}
}

@article{LBMUCJ:2017,
  title = {High-Speed Photonic Reservoir Computing Using a Time-Delay-Based Architecture: Million Words per Second Classification},
  author = {Larger, Laurent and Bayl\'on-Fuentes, Antonio and Martinenghi, Romain and Udaltsov, Vladimir S. and Chembo, Yanne K. and Jacquot, Maxime},
  journal = {Phys. Rev. X},
  volume = {7},
  issue = {1},
  pages = {011015},
  numpages = {14},
  year = {2017},
  month = {Feb},
  publisher = {American Physical Society},
  doi = {10.1103/PhysRevX.7.011015}
}

@article{du2022,
  title={An optoelectronic reservoir computing for temporal information processing},
  author={Du, Wen and Li, Caihong and Huang, Yixuan and Zou, Jihua and Luo, Lingzhi and Teng, Caihong and Kuo, Hao-Chung and Wu, Jiang and Wang, Zhiming},
  journal={IEEE Electron Device Letters},
  volume={43},
  number={3},
  pages={406--409},
  year={2022},
  publisher={IEEE}
}

@article{wang2024,
  title={Harnessing synthetic active particles for physical reservoir computing},
  author={Wang, Xiangzun and Cichos, Frank},
  journal={Nature Communications},
  volume={15},
  number={1},
  pages={774},
  year={2024},
  publisher={Nature Publishing Group UK London}
}

@article{abbas2024,
  title={Classical and Quantum Physical Reservoir Computing for Onboard Artificial Intelligence Systems: A Perspective},
  author={Abbas, AH and Abdel-Ghani, Hend and Maksymov, Ivan S},
  journal={Dynamics},
  volume={4},
  number={3},
  pages={643--670},
  year={2024},
  publisher={MDPI}
}

@article{palacios2024,
  title={Role of coherence in many-body Quantum Reservoir Computing},
  author={Palacios, Ana and Mart{\'\i}nez-Pe{\~n}a, Rodrigo and Soriano, Miguel C and Giorgi, Gian Luca and Zambrini, Roberta},
  journal={Communications Physics},
  volume={7},
  number={1},
  pages={369},
  year={2024},
  publisher={Nature Publishing Group UK London}
}

@misc{Zhu2025,
  doi = {10.48550/ARXIV.2405.04799},
  url = {https://arxiv.org/abs/2405.04799},
  author = {Zhu,  Chuanzhou and Ehlers,  Peter J. and Nurdin,  Hendra I. and Soh,  Daniel},
  title = {Practical Few-Atom Quantum Reservoir Computing},
  publisher = {arXiv},
  year = {2024},
  copyright = {arXiv.org perpetual,  non-exclusive license}
}

@misc{Tran2020,
  doi = {10.48550/ARXIV.2006.08999},
  url = {https://arxiv.org/abs/2006.08999},
  author = {Tran,  Quoc Hoan and Nakajima,  Kohei},
  title = {Higher-Order Quantum Reservoir Computing},
  publisher = {arXiv},
  year = {2020},
  copyright = {arXiv.org perpetual,  non-exclusive license}
}

@article{cucchi2022,
  title={Hands-on reservoir computing: a tutorial for practical implementation},
  author={Cucchi, Matteo and Abreu, Steven and Ciccone, Giuseppe and Brunner, Daniel and Kleemann, Hans},
  journal={Neuromorphic Computing and Engineering},
  volume={2},
  number={3},
  pages={032002},
  year={2022},
  publisher={IOP Publishing}
}

@article{dion2018,
  title={Reservoir computing with a single delay-coupled non-linear mechanical oscillator},
  author={Dion, Guillaume and Mejaouri, Salim and Sylvestre, Julien},
  journal={Journal of Applied Physics},
  volume={124},
  number={15},
  year={2018},
  publisher={AIP Publishing}
}

@article{HSRFG:2015,
  title = {Reservoir computing with a single time-delay autonomous {Boolean} node},
  author = {Haynes, Nicholas D. and Soriano, Miguel C. and Rosin, David P. and Fischer, Ingo and Gauthier, Daniel J.},
  journal = {Phys. Rev. E},
  volume = {91},
  issue = {2},
  pages = {020801},
  numpages = {5},
  year = {2015},
  month = {Feb},
  publisher = {American Physical Society},
  doi = {10.1103/PhysRevE.91.020801}
}

@inproceedings{li2017,
  title={Analog hardware implementation of spike-based delayed feedback reservoir computing system},
  author={Li, Jialing and Zhao, Chenyuan and Hamedani, Kian and Yi, Yang},
  booktitle={2017 International Joint Conference on Neural Networks (IJCNN)},
  pages={3439--3446},
  year={2017},
  organization={IEEE}
}

@article{nichele2017,
  title = {Deep Learning with Cellular Automaton-Based Reservoir Computing},
  volume = {26},
  ISSN = {0891-2513},
  url = {http://dx.doi.org/10.25088/ComplexSystems.26.4.319},
  DOI = {10.25088/complexsystems.26.4.319},
  number = {4},
  journal = {Complex Systems},
  publisher = {Wolfram Research,  Inc.},
  author = {Nichele,  Stefano and Molund,  Andreas},
  year = {2017},
  month = dec,
  pages = {319–340}
}

@inproceedings{goudarzi2013,
  title={DNA reservoir computing: a novel molecular computing approach},
  author={Goudarzi, Alireza and Lakin, Matthew R and Stefanovic, Darko},
  booktitle={International Workshop on DNA-Based Computers},
  pages={76--89},
  year={2013},
  organization={Springer}
}

@article{gonon2024,
  title={Infinite-dimensional reservoir computing},
  author={Gonon, Lukas and Grigoryeva, Lyudmila and Ortega, Juan-Pablo},
  journal={Neural Networks},
  volume={179},
  pages={106486},
  year={2024},
  publisher={Elsevier}
}

@article{Boshgazi2022,
  title = {Virtual reservoir computer using an optical resonator},
  volume = {12},
  ISSN = {2159-3930},
  url = {http://dx.doi.org/10.1364/OME.450256},
  DOI = {10.1364/ome.450256},
  number = {3},
  journal = {Optical Materials Express},
  publisher = {Optica Publishing Group},
  author = {Boshgazi,  Somayeh and Jabbari,  Ali and Mehrany,  Khashayar and Memarian,  Mohammad},
  year = {2022},
  month = feb,
  pages = {1140}
}

@article{Jaurigue2024,
  title = {Chaotic attractor reconstruction using small reservoirs—the influence of topology},
  volume = {5},
  ISSN = {2632-2153},
  url = {http://dx.doi.org/10.1088/2632-2153/ad6ee8},
  DOI = {10.1088/2632-2153/ad6ee8},
  number = {3},
  journal = {Machine Learning: Science and Technology},
  publisher = {IOP Publishing},
  author = {Jaurigue,  Lina},
  year = {2024},
  month = aug,
  pages = {035058}
}

@article{Ma2023,
  title = {A novel approach to minimal reservoir computing},
  volume = {13},
  ISSN = {2045-2322},
  url = {http://dx.doi.org/10.1038/s41598-023-39886-w},
  DOI = {10.1038/s41598-023-39886-w},
  number = {1},
  journal = {Scientific Reports},
  publisher = {Springer Science and Business Media LLC},
  author = {Ma,  Haochun and Prosperino,  Davide and R\"{a}th,  Christoph},
  year = {2023},
  month = aug 
}

@article{He2025,
  title = {Physical reservoir computing on a soft bio-inspired swimmer},
  volume = {181},
  ISSN = {0893-6080},
  url = {http://dx.doi.org/10.1016/j.neunet.2024.106766},
  DOI = {10.1016/j.neunet.2024.106766},
  journal = {Neural Networks},
  publisher = {Elsevier BV},
  author = {He,  Shan and Musgrave,  Patrick},
  year = {2025},
  month = jan,
  pages = {106766}
}

@article{KKGGM:2020,
  title={Dynamical learning of dynamics},
  author={Klos, Christian and Kossio, Yaroslav Felipe Kalle and Goedeke, Sven and Gilra, Aditya and Memmesheimer, Raoul-Martin},
  journal={Phys. Rev. Lett.},
  volume={125},
  number={8},
  pages={088103},
  year={2020},
  publisher={APS}
}

@article{Cucchi2021,
  title = {Reservoir computing with biocompatible organic electrochemical networks for brain-inspired biosignal classification},
  volume = {7},
  ISSN = {2375-2548},
  url = {http://dx.doi.org/10.1126/sciadv.abh0693},
  DOI = {10.1126/sciadv.abh0693},
  number = {34},
  journal = {Science Advances},
  publisher = {American Association for the Advancement of Science (AAAS)},
  author = {Cucchi,  Matteo and Gruener,  Christopher and Petrauskas,  Lautaro and Steiner,  Peter and Tseng,  Hsin and Fischer,  Axel and Penkovsky,  Bogdan and Matthus,  Christian and Birkholz,  Peter and Kleemann,  Hans and Leo,  Karl},
  year = {2021},
  month = aug,
  pages = {eabh0693}
}

@article{Illeperuma2024,
  title = {Novel Directions for Neuromorphic Machine Intelligence Guided by Functional Connectivity: A Review},
  volume = {12},
  ISSN = {2075-1702},
  url = {http://dx.doi.org/10.3390/machines12080574},
  DOI = {10.3390/machines12080574},
  number = {8},
  journal = {Machines},
  publisher = {MDPI AG},
  author = {Illeperuma,  Mindula and Pina,  Rafael and De Silva,  Varuna and Liu,  Xiaolan},
  year = {2024},
  month = aug,
  pages = {574}
}

@article{Zhang2023,
  title = {A Survey on Reservoir Computing and its Interdisciplinary Applications Beyond Traditional Machine Learning},
  volume = {11},
  ISSN = {2169-3536},
  url = {http://dx.doi.org/10.1109/ACCESS.2023.3299296},
  DOI = {10.1109/access.2023.3299296},
  journal = {IEEE Access},
  publisher = {Institute of Electrical and Electronics Engineers (IEEE)},
  author = {Zhang,  Heng and Vargas,  Danilo Vasconcellos},
  year = {2023},
  pages = {81033–81070}
}

@article{Soriano2015,
  title = {Minimal approach to neuro-inspired information processing},
  volume = {9},
  ISSN = {1662-5188},
  url = {http://dx.doi.org/10.3389/fncom.2015.00068},
  DOI = {10.3389/fncom.2015.00068},
  journal = {Frontiers in Computational Neuroscience},
  publisher = {Frontiers Media SA},
  author = {Soriano,  Miguel C. and Brunner,  Daniel and Escalona-MorÃ¡n,  Miguel and Mirasso,  Claudio R. and Fischer,  Ingo},
  year = {2015},
  month = jun 
}

@article{Parlitz2024,
  title = {Learning from the past: reservoir computing using delayed variables},
  volume = {10},
  ISSN = {2297-4687},
  url = {http://dx.doi.org/10.3389/fams.2024.1221051},
  DOI = {10.3389/fams.2024.1221051},
  journal = {Frontiers in Applied Mathematics and Statistics},
  publisher = {Frontiers Media SA},
  author = {Parlitz,  Ulrich},
  year = {2024},
  month = mar 
}

@article{VPPJHBSTEOKP:2020,
  title={Backpropagation algorithms and reservoir computing in recurrent neural networks for the forecasting of complex spatiotemporal dynamics},
  author={Vlachas, Pantelis-Rafail and Pathak, Jaideep and Hunt, Brian R and Sapsis, Themistoklis P and Girvan, Michelle and Ott, Edward and Koumoutsakos, Petros},
  journal={Neural Netw.},
  volume={126},
  pages={191--217},
  year={2020},
  publisher={Elsevier}
}

@article{jaeger2001short,
  title={Short term memory in echo state networks},
  author={Jaeger, Herbert},
  year={2001},
  journal={GMD Forschungszentrum Informationstechnik}
}

@article{carroll2022optimizing,
  title={Optimizing memory in reservoir computers},
  author={Carroll, Thomas L},
  journal={Chaos: An Interdisciplinary Journal of Nonlinear Science},
  volume={32},
  number={2},
  year={2022},
  publisher={AIP Publishing}
}

@article{dambre2012information,
  title={Information processing capacity of dynamical systems},
  author={Dambre, Joni and Verstraeten, David and Schrauwen, Benjamin and Massar, Serge},
  journal={Scientific reports},
  volume={2},
  number={1},
  pages={514},
  year={2012},
  publisher={Nature Publishing Group UK London}
}

@article{ma2023efficient,
  title={Efficient forecasting of chaotic systems with block-diagonal and binary reservoir computing},
  author={Ma, Haochun and Prosperino, Davide and Haluszczynski, Alexander and R{\"a}th, Christoph},
  journal={Chaos: An Interdisciplinary Journal of Nonlinear Science},
  volume={33},
  number={6},
  year={2023},
  publisher={AIP Publishing}
}

@article{rosenstein1993practical,
  title={A practical method for calculating largest Lyapunov exponents from small data sets},
  author={Rosenstein, Michael T and Collins, James J and De Luca, Carlo J},
  journal={Physica D: Nonlinear Phenomena},
  volume={65},
  number={1-2},
  pages={117--134},
  year={1993},
  publisher={Elsevier}
}

@article{zhai2023emergence,
  title={Emergence of a resonance in machine learning},
  author={Zhai, Zheng-Meng and Kong, Ling-Wei and Lai, Ying-Cheng},
  journal={Physical Review Research},
  volume={5},
  number={3},
  pages={033127},
  year={2023},
  publisher={APS}
}

@article{takens2002reconstruction,
  title={The reconstruction theorem for endomorphisms},
  author={Takens, Floris},
  journal={Bulletin of the Brazilian Mathematical Society},
  volume={33},
  number={2},
  pages={231--262},
  year={2002},
  publisher={Springer}
}

@article{wang2014general,
  title={A general stochastic algorithmic framework for minimizing expensive black box objective functions based on surrogate models and sensitivity analysis},
  author={Wang, Yilun and Shoemaker, Christine A},
  journal={arXiv preprint arXiv:1410.6271},
  year={2014}
}

@book{goldberg2013genetic,
  title={Genetic algorithms},
  author={Goldberg, David E},
  year={2013},
  publisher={pearson education India}
}

@article{griffith2019forecasting,
  title={Forecasting chaotic systems with very low connectivity reservoir computers},
  author={Griffith, Aaron and Pomerance, Andrew and Gauthier, Daniel J},
  journal={Chaos: An Interdisciplinary Journal of Nonlinear Science},
  volume={29},
  number={12},
  year={2019},
  publisher={AIP Publishing}
}

@article{KFGL:2021a,
  title = {Machine learning prediction of critical transition and system collapse},
  author = {Kong, Ling-Wei and Fan, Hua-Wei and Grebogi, Celso and Lai, Ying-Cheng},
  journal = {Phys. Rev. Research},
  volume = {3},
  issue = {1},
  pages = {013090},
  numpages = {14},
  year = {2021},
  month = {Jan},
  publisher = {American Physical Society},
  doi = {10.1103/PhysRevResearch.3.013090},
  url = {https://link.aps.org/doi/10.1103/PhysRevResearch.3.013090}
}

@article{KFGL:2021b,
  title = {Emergence of transient chaos and intermittency in machine learning},
  author = {Kong, Ling-Wei and Fan, Hua-Wei and Grebogi, Celso and Lai, Ying-Cheng},
  journal = {J. Phys. Complex.},
  volume = {2},
  issue = {},
  pages = {035014},
  year = {2021},
  url = {https://iopscience.iop.org/article/10.1088/2632-072X/ac0b00}
}

@article{PKMZGHL:2024,
  title = {Machine learning prediction of tipping in complex dynamical systems},
  author = {Panahi, Shirin and Kong, Ling-Wei and Moradi, Mohammadamin and Zhai, Zheng-Meng and Glaz, Bryan and Haile, Mulugeta and Lai, Ying-Cheng},
  journal = {Phys. Rev. Res.},
  volume = {6},
  issue = {4},
  pages = {043194},
  numpages = {18},
  year = {2024},
  month = {Nov},
  publisher = {American Physical Society},
  doi = {10.1103/PhysRevResearch.6.043194},
  url = {https://link.aps.org/doi/10.1103/PhysRevResearch.6.043194}
}

@article{Lorenz:1963,
author={E. N. Lorenz},
title = "Deterministic nonperiodic flow",
year={1963},
journal={J. Atmos. Sci.},
volume={20},
pages = {130-141}  }

@article{HH:1991,
  title={Chaos in a three-species food chain},
  author={Hastings, Alan and Powell, Thomas},
  journal={Ecology},
  volume={72},
  number={3},
  pages={896--903},
  year={1991},
  publisher={Wiley Online Library}
}

@article{MY:1994,
  title={Nonlinear dynamics and population disappearances},
  author={McCann, Kevin and Yodzis, Peter},
  journal={Ame. Naturalist},
  volume={144},
  number={5},
  pages={873--879},
  year={1994},
  publisher={University of Chicago Press}
}

@book{Pardo:book,
  title={Statistical Inference Based on Divergence Measures},
  author={Pardo, L.},
  series={Statistics: A Series of Textbooks and Monographs},
  address = {Boca Raton, Florida},
  year={2018},
  publisher={CRC Press}
}
\clearpage

\newpage
\appendix

\begin{widetext}

\renewcommand{\thesection}{S\arabic{section}}  
\renewcommand{\thetable}{S\arabic{table}}  
\setcounter{table}{0}
\renewcommand{\thefigure}{S\arabic{figure}}
\setcounter{figure}{0}
\renewcommand{\theequation}{S.\arabic{equation}}
\setcounter{equation}{0}

\section*{Supplementary information}

\section{Vestibular System} \label{secA1}

The vestibular system is a sensory mechanism responsible for detecting head movement and maintaining balance~\cite{highstein2004,goldberg2012}. It comprises two primary components: the semicircular canals (SCCs) and the otolith organs, both embedded within the inner ear~\cite{khan2013}. The SCCs detect rotational movements, while the otolith organs sense linear accelerations and gravitational forces~\cite{cullen2019}. Each of these structures terminates in specialized hair cells, which transduce mechanical displacements caused by the movement of the endolymph into electrical signals~\cite{hudspeth1979,guth1998}. These signals are then transmitted to the brain for motion perception, balance control, and spatial orientation, as shown in Fig.~\ref{fig:figsi_v}.

The vestibular system plays an essential role in various biological and engineering applications~\cite{wall2003,moshizi2022}. In biological systems, it integrates multisensory feedback to contribute to postural stability, gaze stabilization, and motor coordination~\cite{lacour1993,whitney2009}. In control and robotics, the principles of vestibular function inspire stabilization strategies for humanoid robots, adaptive control mechanisms, and autonomous vehicle navigation systems~\cite{fransson2003,patane2004,tin2005, mergner2009}. Additionally, modeling vestibular dysfunction has applications in medical diagnostics, rehabilitation, and prosthetic development, particularly for individuals with balance disorders~\cite{whitney2016,sulway2019}.

The development of mathematical models for the vestibular system and the understanding of its functional dynamics have been a central focus of research due to its critical role in motion perception and control~\cite{santos2017,Momani2018}. Early models primarily focused on two-dimensional linear approximations for the SCCs and otolith organs to capture their fundamental dynamics. For example, the foundation of mathematical modeling in the vestibular system can be traced back to Steinhausen’s studies (1927–1933), where a linear second-order model was introduced to describe SCC dynamics~\cite{steinhausen1933, straka2021}. These models aimed to explain the characteristics of vestibular-induced eye movements in fish (pike) and laid the groundwork for subsequent research. Over time, these models have been extensively refined and expanded, incorporating additional physiological and biomechanical factors to enhance their predictive capabilities. By 1949 and 1956, this model had been further developed for human subjects through experiments based on neurophysiological and mechanical principles~\cite{gernandt1949,vanEgmond1949,groen1956,groen1957}. A torsion-pendulum model was proposed, incorporating the friction of the endolymphatic fluid, the moment of inertia of the canals, and the stiffness and elastic properties of the cupula~\cite{vanEgmond1949}. Results demonstrated that this model proved capable of predicting the cupula’s position for various head movements, including constant acceleration, sinusoidal oscillations, and step changes in angular velocity.

While the torsion-pendulum model effectively describes angular velocity sensing in SCCs, it does not fully capture rotational sensation. To address these limitations, researchers introduced biocybernetic techniques inspired by biological control systems, leading to models that included quantitative descriptions of adaptation~\cite{young1969}. In 1971, mathematical formulations were further expanded to explain nystagmus patterns and general vestibular phenomena, such as adaptation and secondary nystagmus~\cite{schmid1971}. Using experimental data from young healthy subjects, researchers quantified key model parameters to enhance accuracy. To provide a broader understanding of vestibular processing beyond individual components, a systems perspective of vestibular organs was proposed in 1974, focusing on the primary functions and integrated operations of the vestibular system~\cite{benson1974}.

As research progressed, the engineering significance of the vestibular system was increasingly explored. For instance, a 1966 study provided a comprehensive framework by analyzing the vestibular system's role in human spatial orientation and manual vehicle control~\cite{meiry1965}. This work introduced control models for vestibular sensors, examined eye stabilization subsystems, and investigated the impact of motion cues on closed-loop manual control mechanisms. In 2005, a major advancement in vestibular modeling occurred with the development of motion cueing algorithms. These algorithms transformed simulated aircraft dynamics into motion commands, incorporating human vestibular models to improve pilot training and simulation accuracy~\cite{houck2005}. Since then, research has continued to refine vestibular models for diverse applications, including robotics, virtual reality, neuroengineering, and biomedical devices~\cite{peters1969,rabbitt2004,gastaldi2009,stewart2009,bradshaw2010,mccollum2010,asadnia2016,asadi2017,Momani2018,paulin2019,ramat2019,lu2021,minyailo2024}.

\begin{figure}
\centering
\includegraphics[width=0.6\linewidth]{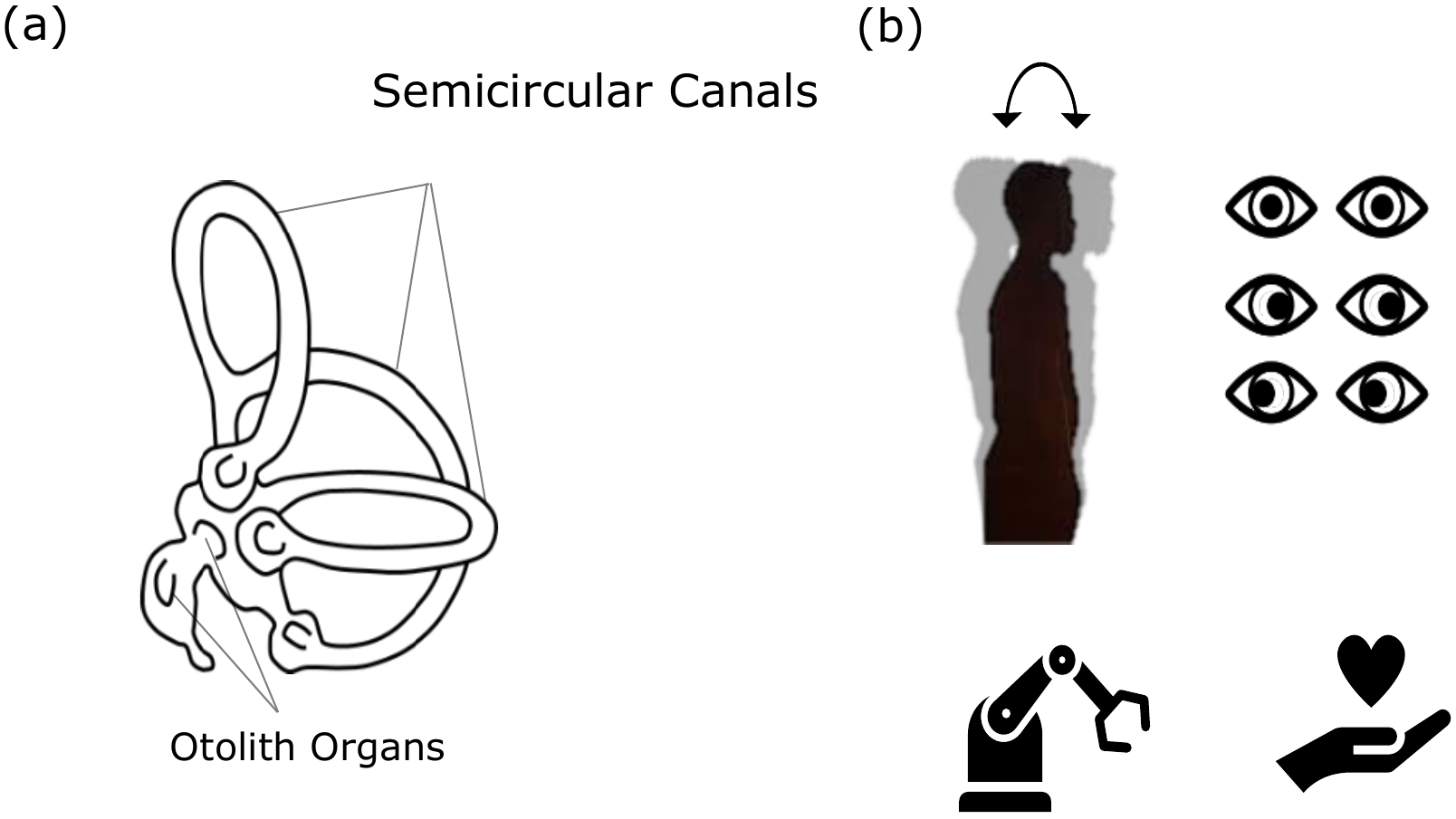}
\caption{Schematic representation of the vestibular system and its functionality. (a) Anatomical schematic of the vestibular system. (b) Overview of functional applications, including postural stability, motion control, eye movement stabilization, robotics, and autonomous systems.}
\label{fig:figsi_v}
\end{figure}

Contemporary mathematical frameworks for understanding the biomechanical and neurophysiological functions of the vestibular system typically incorporate both the SCCs and the otolith organs, alongside the role of hair cells, which function as transducers or biosensors converting mechanical movement into electrical signals~\cite{moore2004,green2010,glasauer2018,clark2019,paulin2019}. These models generally consist of two primary components: a linear second-order system representing SCC dynamics and a mathematical neuron model, such as the Hodgkin-Huxley model, to describe neural signal processing~\cite{alexandrov2007,sadovnichy2007,aleksandrov2014,canelo2018}. The SCC model captures the fluid dynamics within the canals, where endolymph movement induces cupula displacement, a response approximated by a damped second-order differential equation. Conversely, the Hodgkin-Huxley model (or similar neuronal models) simulates the ionic mechanisms of hair cells, transforming mechanical deflections into voltage changes that encode vestibular information.

\section{Hyperparameters}\label{secH}

The reservoir computer contains several hyperparameters that require optimization to achieve desired performance. In the proposed vestibular reservoir computer, these hyperparameters include the range $[-\gamma,\gamma]$ for the elements of the input matrix $\mathbb{W}_{\rm in}$, the spectral radius $\rho$ of the recurrent network in the hidden layer, the regularization coefficient $\lambda$, and the sparsity measure $d$, which determines the connectivity of the network matrix $\mathbb{A}$. These hyperparameters define the reservoir architecture and significantly impact its performance.

Two commonly employed methods for hyperparameter optimization are random search and Bayesian optimization. In the random search method, hyperparameters are selected randomly from a predefined domain. Specifically, given vector time-series input data, the training phase yields the output matrix $\mathbb{W}_{\rm out}$. During the validation phase, the reservoir continues to receive the input signal, but the output matrix remains fixed to generate one-step predictions. The average one-step prediction error is then calculated. If this error is unsatisfactory (e.g., exceeding a predefined threshold), a new set of hyperparameters is selected. This iterative process comprising training and validation continues until the error meets the required criteria. Bayesian optimization~\cite{griffith2019forecasting} offers a data-efficient alternative, identifying optimal hyperparameters by probabilistically modeling the performance landscape to select which configurations to test next. Other methods for hyperparameter optimization include surrogate models~\cite{wang2014general} and genetic algorithms~\cite{goldberg2013genetic}.

The optimized hyperparameters selected for the vestibular reservoir computer for the Lorenz and food-chain inputs are listed in Tab.~\ref{tab:example}.

\begin{table}[h!]
\caption{Hpyerparameters of the coupled and uncoupled vestibular reservoir computer for Lorenz and chaotic food-chain systems.}	\label{tab:example}
\begin{tabular}{|c|c|c|c|c|c|}
\hline
Physical reservoir computer&	Model & $\gamma$ & $\rho$ & $\lambda$ & $d$\\ \hline
Coupled    &	Lorenz  & 1 &   0.8 & $10^{-4}$ &0.4 \\ \hline
	$\cdots$   &    food-chain   & 3.5  &  0.2 & $10^{-5}$ &0.4   \\ \hline
Uncoupled & 	Lorenz   & 1 &    0.5 &$10^{-4}$ &0.4   \\ \hline
	$\cdots$  & food-chain  & 3.5 &    0.7 & $10^{-5}$ &0.4  \\ \hline
\end{tabular}
\end{table}

\section{Predictive statistics} \label{si_statistic}

{\em Performance Evaluation}. To evaluate the short-term prediction accuracy of the vestibular reservoir computer, we analyze the training and validation accuracy to quantify the model's fidelity to the ground truth over a finite interval. Specifically, the predicted trajectory $\hat{\mathbf{y}}(t)$ is compared with the actual system trajectory $\mathbf{y}(t)$. Performance is evaluated using the normalized root-mean-square error (NRMSE), defined as: 
\begin{align}
\text{NRMSE} = \frac{\sqrt{\frac{1}{T}\sum_{t=1}^{T} (y(t) - \hat{y}(t))^2}}{\sigma_y},
\end{align} 
where $\sigma_y$ is the standard deviation of the true output. A lower NRMSE indicates superior predictive performance.

We also assess the long-term stability of both coupled and uncoupled vestibular reservoir computers, specifically, their ability to maintain coherent trajectories without divergence. Long-term predictive capability is evaluated by measuring the deviation of the reconstructed attractor from the ground-truth attractor using statistical metrics such as the largest Lyapunov exponent, the deviation value (DV), and the Kullback-Leibler (KL) divergence.

The largest Lyapunov exponent quantifies the average exponential divergence of nearby trajectories in state space and is calculated directly from the time series using the standard delay-coordinate embedding method~\cite{takens2002reconstruction,rosenstein1993practical}. A positive largest Lyapunov exponent signifies sensitivity to initial conditions, a hallmark of chaos. Conversely, a value close to zero indicates that trajectories remain predictable over long horizons, implying the system neither diverges chaotically nor collapses to a fixed point.

The deviation value (DV)\cite{zhai2023emergence} and KL divergence\cite{Pardo:book} measure the distance between probability distributions; the latter specifically quantifies the information loss when approximating the true distribution with the predicted one. To compute these metrics, high-dimensional attractors are projected onto a two-dimensional plane overlaid with a uniform grid. We calculate the visitation frequency for each grid cell $(i,j)$ for both the predicted and ground-truth trajectories over the interval $0 \le t \le T_{\rm pred}$. Let $\hat{f}_{ij}$ and $f_{ij}$ denote the visitation frequencies for the predicted and ground-truth attractors, respectively. The DV is calculated as: 
\begin{align} \label{eq:C5_Pred_DV} 
	{\rm DV} = \sum_{i}\sum_{j} \sqrt{(f_{ij} - \hat{f}_{ij})^2}, 
\end{align} 
where the summation extends over all grid cells. Likewise, the KL divergence is given by~\cite{Pardo:book}: 
\begin{align} \label{eq:C5_Pred_KL} 
{\rm KL} = \sum_{i}\sum_{j} f_{ij} \log{\left(\frac{f_{ij}}{\hat{f}_{ij}}\right)}. 
\end{align} 
Smaller values for both DV and KL indicate a better statistical match between the predicted and true attractors.

\section{Performance of uncoupled vestibular reservoir computer}

In this section, we present additional results demonstrating that the proposed uncoupled vestibular reservoir computer is capable of achieving accuracy competitive with its coupled counterpart, provided they share the same eigenvalue spectrum.

\begin{figure} [ht!]
\centering
\includegraphics[width=0.7\linewidth]{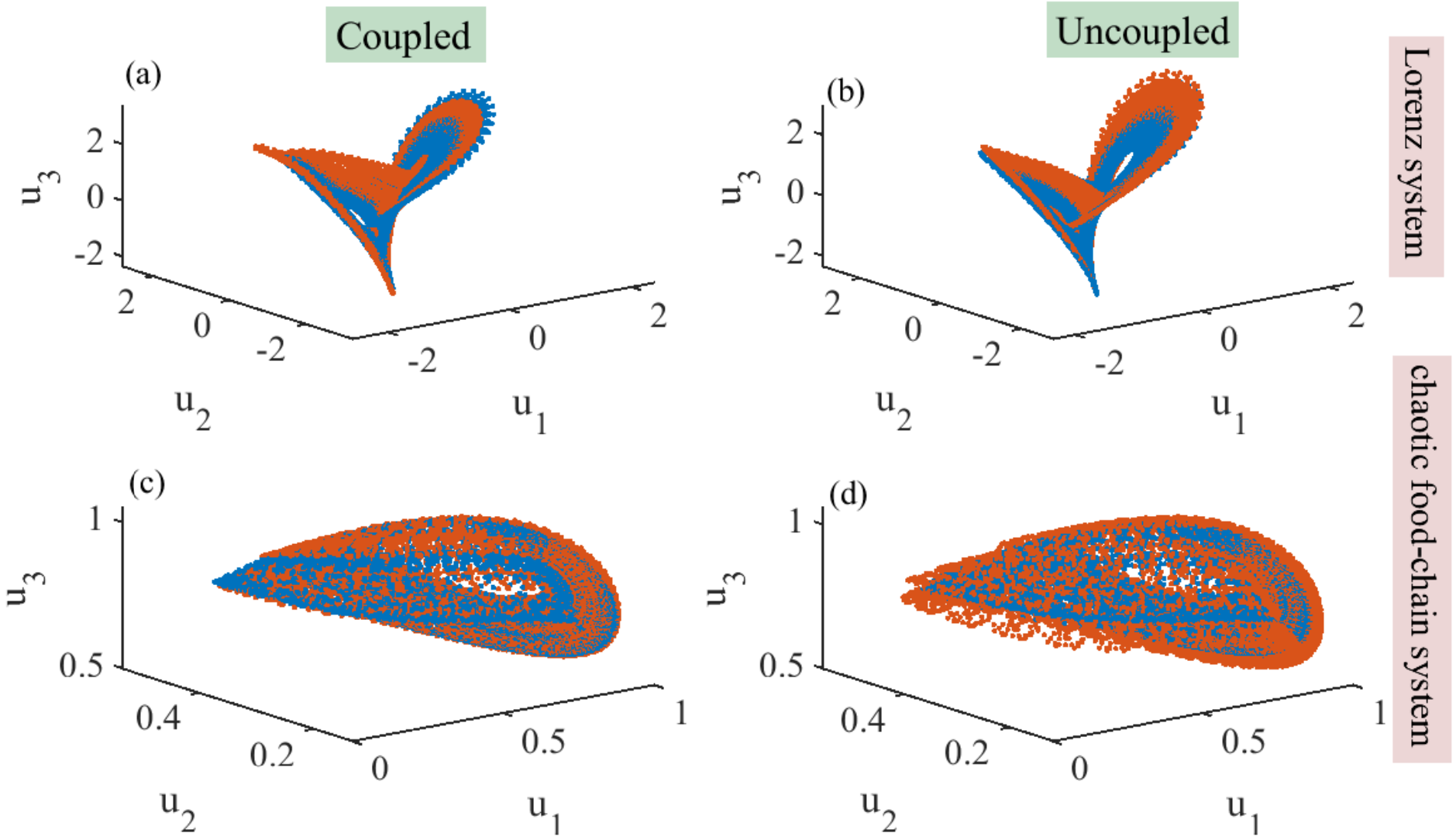}
\caption{Performance of the vestibular reservoir computer: Closed-loop attractor reconstruction. Comparison of ground truth (blue) and predicted (orange) attractors for all state variables of the Lorenz and chaotic food-chain systems. Results are shown for the (a-c) coupled and (d-f) uncoupled vestibular reservoir configurations. The autonomous reservoir successfully reproduces the underlying chaotic dynamics, validating its ability to capture the long-term behavior of the original systems. Simulation parameters are identical to those used in Fig.~\ref{fig:fig2} of the main text.}
\label{fig:figsi31}
\end{figure}

\begin{figure}[ht!]
\centering
\includegraphics[width=0.7\linewidth]{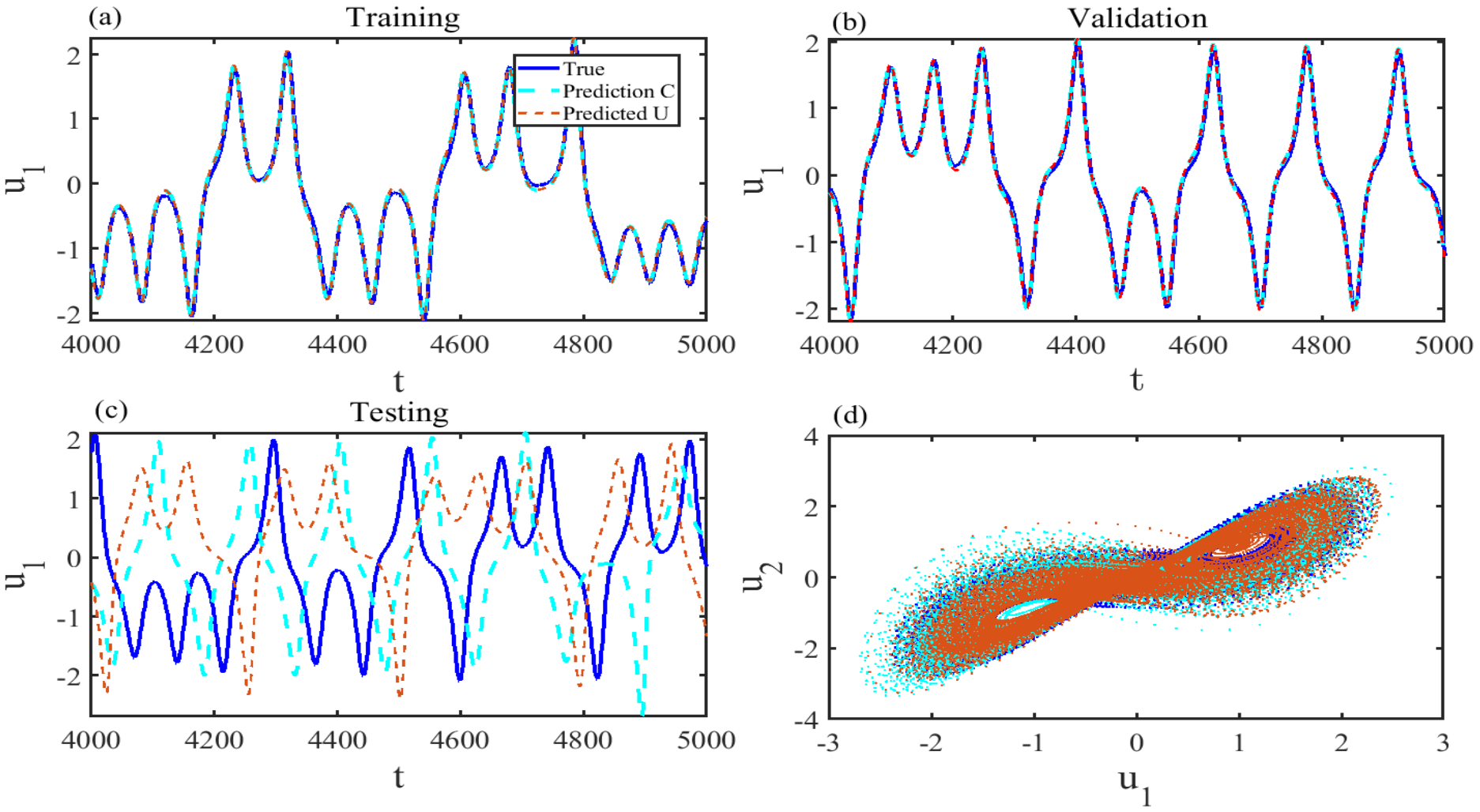}
\caption{Performance of the vestibular reservoir computer for the Lorenz system. (a) Training phase time series showing the ground truth (blue) alongside predicted outputs for the coupled (cyan) and uncoupled configurations. Note that the uncoupled network is configured to share the same eigenvalue spectrum as the coupled one. (b) Validation phase performance, evaluating the trained reservoir's generalization accuracy on unseen data in an open-loop setup. (c-d) Testing phase time series and closed-loop attractor reconstruction. The autonomous reservoir successfully reproduces the underlying chaotic dynamics, confirming its ability to capture long-term system behavior. Both coupled and uncoupled configurations exhibit comparable performance.}
\label{fig:figsi2}
\end{figure}

\begin{table} [h!]
\caption{Comparison of mean predictive statistics for the Lorenz system. Results represent an ensemble average over 50 realizations for coupled and uncoupled (sharing the same eigenvalue spectrum) vestibular reservoir computers with $N=30$ nodes. The theoretical largest Lyapunov exponent for the system is approximately $0.03$.}
\label{tab:tabS41}
	\begin{tabular}{|c|c|c|}
		\hline
		Statistics & Coupled & Uncoupled with same eigenvalue \\ \hline
		Training error & 0.0111  &  0.0115 \\ \hline
		Validation error  & 0.0114  & 0.0117 \\ \hline
		Deviation value  & 0.331  & 0.352  \\ \hline
		KL Divergence  &  0.0005 & 0.0009\\ \hline
		Largest Lyapunov exponent (predicted)  &  0.0308& 0.0307  \\ \hline
	\end{tabular}
\end{table}

\begin{figure}[ht!]
\centering
\includegraphics[width=0.7\linewidth]{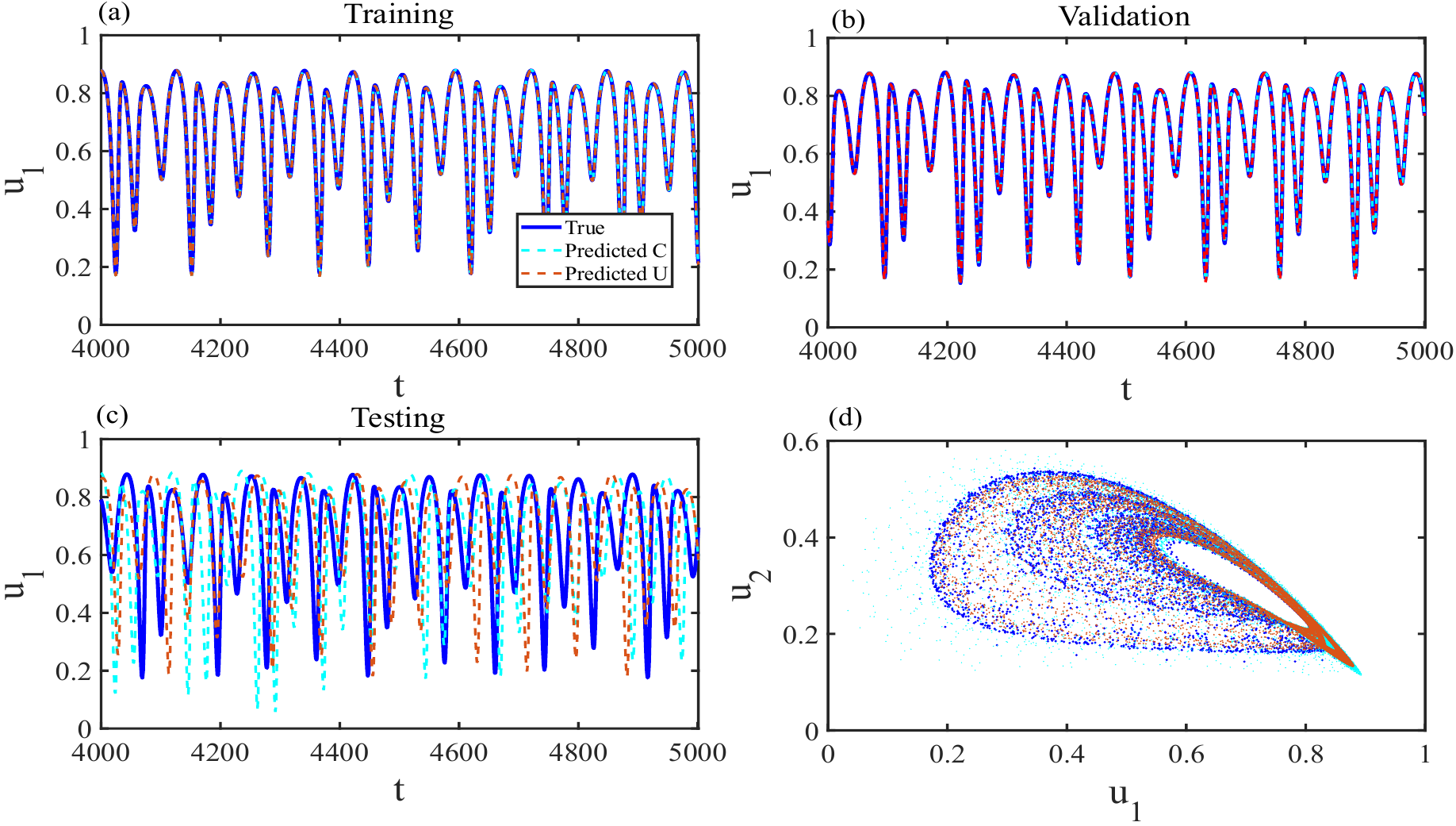}
\caption{Performance of the vestibular reservoir computer for the chaotic food-chain system. (a) Training phase time series displaying the ground truth (blue) alongside predicted outputs for the coupled (cyan) and uncoupled (red) configurations, where the uncoupled network shares the same eigenvalue spectrum as the coupled one. (b) Validation phase performance, assessing generalization accuracy on unseen data in an open-loop setup. (c-d) Testing phase time series and closed-loop attractor reconstruction. The autonomous reservoir successfully reproduces the underlying chaotic dynamics, confirming its ability to capture long-term system behavior. Both configurations exhibit comparable performance.}
\label{fig:figsi3}
\end{figure}

\begin{table}[h!]
\caption{Comparison of mean predictive statistics for the chaotic food-chain system. Results represent an ensemble average over 50 realizations for coupled and uncoupled (sharing the same eigenvalue spectrum) vestibular reservoir computers with $N=30$ nodes. The theoretical largest Lyapunov exponent for the system is approximately $0.021$.} \label{tab:tabS42}
\begin{tabular}{|c|c|c|}
		\hline
		Statistics & Coupled & Uncoupled with same eigenvalue \\ \hline
		Training error & 0.009 &  0.0098 \\ \hline
		Validation error  & 0.0095  & 0.0099 \\ \hline
		Deviation value  & 0.34  & 0.375 \\ \hline
		KL Divergence  &  0.0005 & 0.0008\\ \hline
		Largest Lyapunov exponent (predicted)  &  0.0237 & 0.0225  \\ \hline
\end{tabular}
\end{table}

\section{Memory function and memory capacity of a linear reservoir Computer} \label{SI:Sec5}

The reservoir state of a linear reservoir computer with $N$ nodes is given by
\begin{align}
\mathbf{r}(t+1) = \mathbb{A}\cdot\mathbf{r}(t) + \mathbf{W}_{\rm in}u(t),
\end{align}
where $u(t) \in \mathcal{R}$ is a stochastic input, $\mathbb{A}\in \mathcal{R}^{N \times N}$ is the internal weight matrix, and $\mathbf{W}_{\rm in} \in \mathcal{R}^{N}$ is the input weight vector. The memory function is defined as
\begin{align}
{\rm M}_F[\mathbb{R},\mathbf{u}_{\tau}]=\frac{\mathbf{u}_{\tau}\cdot\mathbb{R}^{\intercal}\cdot(\mathbb{R}\cdot\mathbb{R}^{\intercal})^{-1}\cdot\mathbb{R}\cdot\mathbf{u}_{\tau}^{\intercal}}{\mathbf{u}_{\tau}\cdot\mathbf{u}_{\tau}^{\intercal}}
\end{align}
where $\mathbb{R}$ is the reservoir state matrix and $\mathbf{u}_{\tau}$ is the vector of delayed time series of past input. The reservoir state $\mathbf{r}(t)$ can be obtained iteratively as
\begin{align}
\mathbf{r}(t+1) = \mathbb{A}^{t+1}\cdot \mathbf{r}(0) + \mathbb{A}^{t}\cdot(\mathbf{W}_{in} u(1)) + \dots + \mathbb{A}(\mathbf{W}_{in} u(t)) + \mathbf{W}_{in}u(t),
\end{align}
or equivalently,
\begin{align}
\mathbf{r}(t+1) = \mathbb{A}^{t+1}\cdot \mathbf{r}(0) + \sum_{i=1}^{t+1} \mathbb{A}^{i-1}\cdot \mathbf{W}_{in} u(t-i+1).
\end{align}
For the initial state $\mathbf{r}(0) =\mathbf{0}$, we have
\begin{align}
\label{eq:eq18}
\mathbf{r}(t+1) = \sum_{i=1}^{t+1} \mathbb{A}^{i-1}\cdot\mathbf{W}_{in} u(t-i+1).
\end{align}
To simplify the expression of the reservoir state, we exploit the eigenvalue decomposition of the internal weight matrix $\mathbb{A}$:
\begin{align} \label{eq:eq19}
\mathbb{A} = \mathbb{P}\cdot\Sigma\cdot\mathbb{P}^{-1},
\end{align}
where $\Sigma = \mathrm{diag}(\lambda_1, \dots, \lambda_m, \dots, \lambda_N)$ with $\lambda_m$ being the $m$-th eigenvalue of $\mathbb{A}$, $\Sigma$ and $\mathbb{P}$ being $N \times N$ matrices. Substituting Eq.~\eqref{eq:eq19} in Eq.~\eqref{eq:eq18}, we get
\begin{align}\label{eqSN}
\mathbf{r}(t+1) = \sum_{i=1}^{t+1} \mathbb{P}\cdot{\Sigma}^{i-1}\cdot \mathbb{P}^{-1}\cdot\mathbf{W}_{in} u(t-i+1).
\end{align}
where, $\Sigma^{i-1}$ stands for the $(i-1)$-th power of the matrix $\Sigma$. Letting
\begin{align*}
\mathbb{P}^{-1}\cdot \mathbf{W}_{in} = 
\begin{pmatrix}
\mathbf{p}_1^{\intercal}\cdot\mathbf{W}_{in} \\[6pt]
\vdots \\[6pt]
\mathbf{p}_N^{\intercal}\cdot\mathbf{W}_{in}
\end{pmatrix},
\end{align*}
where $\mathbf{p}_i^{\intercal}$ is the $i$-th row vector of $\mathbb{P}^{-1}$ and substituting it in Eq.~\eqref{eqSN}, we obtain
\begin{align} \label{eq:eq20}
\mathbf{r}(t+1) = \mathbb{P}\cdot \sum_{i=1}^{t+1}
\begin{pmatrix}
(\mathbf{p}_1^{\intercal}\cdot\mathbf{W}_{in})\,\lambda_1^{i-1} \\
\vdots \\
(\mathbf{p}_N^{\intercal} \cdot\mathbf{W}_{in})\,\lambda_N^{i-1}
\end{pmatrix}
u(t-i+1).
\end{align}
Expanding Eq.~\eqref{eq:eq20}, we obtain 
\begin{align*}
\mathbf{r}(t+1)= \mathbb{P}_{N \times N}\cdot
\begin{pmatrix}
(\mathbf{p}_1^{\intercal}\cdot \mathbf{W}_{in})\lambda_1^{0} u(t) + (\mathbf{p}_1^{\intercal}\cdot \mathbf{W}_{in})\lambda_1^{1} u(t-1) + \cdots + (\mathbf{p}_1^{\intercal}\cdot \mathbf{W}_{in})\lambda_1^{t} u(0) \\[6pt]
(\mathbf{p}_2^{\intercal}\cdot \mathbf{W}_{in})\lambda_2^{0} u(t) + (\mathbf{p}_2^{\intercal}\cdot \mathbf{W}_{in})\lambda_2^{1} u(t-1) + \cdots + (\mathbf{p}_2^{\intercal}\cdot \mathbf{W}_{in})\lambda_2^{t} u(0) \\[6pt]
\vdots \\[6pt]
(\mathbf{p}_N^{\intercal}\cdot \mathbf{W}_{in})\lambda_N^{0} u(t) + (\mathbf{p}_N^{\intercal}\cdot \mathbf{W}_{in})\lambda_N^{1} u(t-1) + \cdots + (\mathbf{p}_N^{\intercal}\cdot \mathbf{W}_{in})\lambda_N^{t} u(0)
\end{pmatrix}_{N \times 1}.
\end{align*}
Defining
\begin{equation*}
\mathbf{\Lambda}_k = 
\begin{pmatrix}
\lambda_k^{T-1} \\
\lambda_k^{T-2} \\
\vdots \\
\lambda_k^{0}
\end{pmatrix},
\mathbf{u}_\tau = 
\begin{pmatrix}
u(t-\tau-(T-1)) \\
u(t-\tau-(T-2))\\
\vdots \\
u(t-\tau-0)
\end{pmatrix},
\end{equation*}
where $\mathbf{u}_{\tau}$ is the input time series, shifted by $\tau$ steps into the past over a large time window $T$,
we obtain
\begin{align} \label{eq:eqx}
\mathbf{r}(t+1-\tau) = \mathbb{Q}\cdot
\begin{pmatrix}
\mathbf{\Lambda}_1^{\intercal}\cdot\mathbf{u}_\tau \\
\vdots \\
\mathbf{\Lambda}_N^{\intercal}\cdot\mathbf{u}_\tau
\end{pmatrix},
\end{align}
where $\tau$ assumens values $ 0, 1, 2, \dots, T-1$, $t=\tau,\dots,T-1$ and $T$ is the length of the input signal. The $N\times N$-dimensional matrix $\mathbb{Q}$ is the product of the $i$-th column of $\mathbb{P}$ with $\mathbf{p}_{i}^{\intercal}\cdot\mathbf{W}_{in}$, and each $\mathbf{\Lambda}_k^{\intercal}\cdot\mathbf{u}_{\tau}$ is a scalar for $k = 1, 2, \dots, N$. Equation~\eqref{eq:eqx} can then be written as
\begin{align}
\mathbb{R}= \mathbb{Q}\cdot\mathbb{Z}^{\intercal}, 
\end{align}
where
\begin{equation*} \label{eqZ}
\mathbb{Z}^T = 
\begin{pmatrix}
\mathbf{\Lambda}_1^{\intercal}\cdot \mathbf{u}_{T-1} & \cdots & \mathbf{\Lambda}_1^{\intercal}\cdot\mathbf{u}_{1} &\mathbf{\Lambda}_1^{\intercal}\cdot\mathbf{u}_{0} \\
\vdots \\
\mathbf{\Lambda}_N^{\intercal}\cdot\mathbf{u}_{T-1} & \cdots & \mathbf{\Lambda}_N^{\intercal}\cdot \mathbf{u}_{1} &\mathbf{\Lambda}_N^{\intercal}\cdot \mathbf{u}_{0}
\end{pmatrix},
\end{equation*}
and
\begin{align}
\label{eq:eq21}
\mathbf{u}_\tau\cdot\mathbb{R}^{\intercal}\cdot(\mathbb{R}\cdot\mathbb{R}^{\intercal})^{-1}\cdot \mathbb{R} \cdot\mathbf{u}_{\tau}^{\intercal}
= (\mathbb{R}\cdot \mathbf{u}_{\tau}^{\intercal})^{\intercal} \cdot(\mathbb{R}\cdot \mathbb{R}^{\intercal})^{-1}\cdot (\mathbb{R}\cdot \mathbf{u}_{\tau}^{\intercal}).
\end{align}
Replacing $\mathbb{R}= \mathbb{Q}\cdot\mathbb{Z}^T$ and using matrix identity [$(A\cdot B)^{-1}=B^{-1}\cdot {A}^{-1}$] in Eq.~\eqref{eq:eq21}, we get
\begin{align*}
&=(\mathbb{Q}\cdot \mathbb{Z}^{\intercal} \cdot\mathbf{u}_{\tau}^{\intercal})^{\intercal}\cdot \, \big( \mathbb{Q}\cdot \mathbb{Z}^{\intercal} (\mathbb{Q}\cdot \mathbb{Z}^{\intercal})^{\intercal} \big)^{-1}\cdot \, (\mathbb{Q} \mathbb{Z}^{\intercal}\cdot \mathbf{u}_{\tau}^{\intercal}),
\\
&=(\mathbb{Z}^{\intercal}\cdot \mathbf{u}_{\tau}^{\intercal})^{\intercal} \cdot\mathbb{Q}^{\intercal}\cdot \, (\mathbb{Q}\cdot \mathbb{Z}^{\intercal}\cdot \mathbb{Z}\cdot \mathbb{Q}^{\intercal})^{-1}\cdot\, (\mathbb{Q} \cdot\mathbb{Z}^{\intercal}\cdot\mathbf{u}_{\tau}^{\intercal}),
\\
&= (\mathbb{Z}^{\intercal}\cdot\mathbf{u}_{\tau}^{\intercal})^{\intercal}\cdot \mathbb{Q}^{\intercal} \cdot(\mathbb{Q}^{\intercal})^{-1}\cdot (\mathbb{Z}^{\intercal}\cdot \mathbb{Z})^{-1}\cdot \mathbb{Q}^{-1}\cdot \mathbb{Q}\cdot \mathbb{Z}^{\intercal}\cdot \mathbf{u}_{\tau}^{\intercal},
\\
&=(\mathbb{Z}^{\intercal}\cdot \mathbf{u}_{\tau}^{\intercal})^{\intercal}\cdot(\mathbb{Z}^{\intercal}\cdot \mathbb{Z})^{-1}\cdot(\mathbb{Z}^{\intercal}\cdot\mathbf{u}_{\tau}^{\intercal}).
\end{align*}
Equation~\eqref{eq:eq3} becomes
\begin{align}
\rm{M}_F(\mathbb{R},\mathbf{u}_{\tau}) 
= \frac{(\mathbb{Z}^{\intercal}\cdot\mathbf{u}_{\tau}^{\intercal})^{\intercal}\cdot(\mathbb{Z}^{\intercal}\cdot \mathbb{Z})^{-1}\cdot(\mathbb{Z}^{\intercal}\cdot\mathbf{u}_{\tau}^{\intercal})}{ \mathbf{u}_\tau\cdot \mathbf{u}_{\tau}^{\intercal}}.
\end{align}
Further, the entries of $\mathbb{Z}^{\intercal}$ are given by:
\begin{align}
[z]_k = \sum_{i \geq 1} \lambda_k^{i-1} \, u(t-i+1),
\quad \text{where } k = 1,2,\ldots,N \text{ and } i = 1,2,\ldots,T.
\end{align}
When we augment the reservoir state matrix by $\mathbb{R} = [\mathbb {R}_{\text{int}}, \mathbb{R}_s]$, where $\mathbb{R}_s$ has the same dimension as $\mathbb{R}_{\text{int}}$ with entries corresponding to the squared values of $\mathbb{R}_{\text{int}}$, the functional form of all subsequent operations remains unchanged. The only modification is that the effective state dimension $N$ is replaced by $2N$.

Considering stochastic inputs and applying the central limit theorem ($T$ being sufficiently large), the entries of the $N\times T$-dimensional matrix $\mathbb{Z}^{\intercal}\cdot\mathbb{Z}$ in Eq.~\eqref{eqZ}) is given by
\begin{align}
[\mathbb{Z}^{\intercal}\cdot\mathbb{Z}]_{kn} = T \sigma_{u}^{2} \sum_{i \geq 1} (\lambda_{k} \lambda_{n})^{i-1},
\end{align}
where $i = 1, \ldots, T$, $\sigma_{u}$ is the standard deviation of the stochastic input. We have 
\begin{align}
\mathbb{Z}^{\intercal}\cdot\mathbb{Z} = T \sigma_{u}^{2} \mathbb{H}\cdot\mathbb{H}^{\intercal},
\end{align}
where $\mathbb{H}$ is the matrix of eigenvalues of dimension $N \times T$ given by
\begin{align}
\mathbb{H} =
\begin{bmatrix}
\lambda_{1}^{T-1} & \lambda_{1}^{T-2} & \cdots & \lambda_{1} & 1 \\
\vdots & \vdots & \ddots & \vdots & \vdots \\
\lambda_{N}^{T-1} & \lambda_{N}^{T-2} & \cdots & \lambda_{N} & 1
\end{bmatrix},
\end{align}
and \begin{align}
\mathbf{H}_{\tau} = \begin{bmatrix} 
\lambda_{1}^{\tau}, \ldots, \lambda_{N}^{\tau} 
\end{bmatrix}^{\intercal}.
\end{align}
Similarly,
\begin{align}
\mathbb{Z}^{\intercal} \mathbf{u}_{\tau}^{\intercal} \approx T \sigma_{u}^{2} \mathbf{H}_{\tau}.
\end{align}
We have
\begin{align}
\mathbf{u}_{\tau}\cdot\mathbf{u}_{\tau}^{\intercal} \;\approx\; T \sigma_{u}^{2}.
\end{align}
Substituting these values in Eq.~\eqref{eq:eq3}, we get
\begin{align}
\rm{M}_F[\mathbb{R}, \mathbf{u}_{\tau}] 
= \frac{ \big( T \sigma_{u}^{2} \mathbf{H}_{\tau} \big)^{\intercal} \cdot
	\big( T \sigma_{u}^{2} \mathbb{H}\cdot \mathbb{H}^{\intercal} \big)^{-1} \cdot
	\big( T \sigma_{u}^{2} \mathbf{H}_{\tau} \big) }
{ T \sigma_{u}^{2} }.
\end{align}
Simplifying the expression, we obtain the required formula for memory function of a linear reservoir
\begin{align}
\rm{M}_F[\mathbb{R}, \mathbf{u}_{\tau}]
= \mathbf{H}_{\tau}^{\intercal}\cdot (\mathbb{H} \cdot\mathbb{H}^{\intercal})^{-1}\cdot\mathbf{H}_{\tau}.
\end{align}
The memory capacity is given by
\begin{align}
\rm{M}_C = \sum_{\tau = 0}^{T-1} \rm{M}_F[\mathbb{R}, \mathbf{u}_{\tau}]
= \mathrm{tr} \left[ \mathbb{H}^{\intercal}\cdot (\mathbb{H}\cdot\mathbb{H}^{\intercal})^{-1}\cdot\mathbb{H} \right].
\end{align}

\section{Theoretical bound of memory capacity}

For both coupled and uncoupled linear reservoir computers, the theoretical bound on the memory capacity in the limit $T \to \infty$ is given by
\begin{align}
\rm{M}_C = \mathrm{tr} \left[ (\mathbb{H}\cdot \mathbb{H}^{T})^{-1} \cdot\mathbb{H}\cdot \mathbb{H}^{T} \right] = \mathrm{tr}[\mathbb{I}_{N}] = N,
\end{align}
provided $\mathrm{rank}(\mathbb{H}\cdot \mathbb{H}^{T}) = N$. If $\mathbb{H}$ does not have full rank, we can compute the pseudoinverse to obtain 
\begin{align}
\rm{M}_C = \mathrm{rank}(\mathbb{H}) \;\leq N.
\end{align}
The upper bound for $\rm{M}_C$ in linear reservoir is the size of the reservoir $N$. It is worth noting that this theoretical bound holds only for a linear reservoir computer with stochastic input. Currently, no theoretical bounds have been found for nonlinear reservoirs owing to the inherent challenges involved.

\end{widetext}

\end{document}